\pdfoutput=1

\documentclass[11pt]{article}

\usepackage[final]{acl}

\usepackage{times}
\usepackage{latexsym}

\usepackage[T1]{fontenc}

\usepackage[utf8]{inputenc}

\usepackage{microtype}

\usepackage{inconsolata}

\usepackage{graphicx}

\usepackage{amsmath}
\usepackage{todonotes}
\usepackage{booktabs}
\usepackage{subcaption}

\usepackage[commandnameprefix=ifneeded, todonotes={textsize=small}]{changes}
\definechangesauthor[color=teal]{df}
\definechangesauthor[color=blue]{mn}
\definechangesauthor[color=olive]{mg}

\definechangesauthor[color=violet]{lc}

\newcommand{\df}[1]{\textcolor{black}{#1}}

\title{The Unheard Alternative: \\ Contrastive Explanations for Speech-to-Text Models}

\author{
Lina Conti$^{\diamondsuit\spadesuit}$, Dennis Fucci$^{\diamondsuit\spadesuit}$, Marco Gaido$^\spadesuit$, Matteo Negri$^\spadesuit$, \\
\textbf{Guillaume Wisniewski}$^\clubsuit$, \textbf{Luisa Bentivogli}$^\spadesuit$ \\
 $^\diamondsuit$University of Trento, Italy \\
 $^\spadesuit$Fondazione Bruno Kessler, Italy \\
 $^\clubsuit$Laboratoire de Linguistique Formelle, Université Paris Cité, CNRS, Paris, France \\
\texttt{\{lvarellaconti,dfucci,mgaido,negri,bentivo\}@fbk.eu, guillaume.wisniewski@u-paris.fr}
}

\begin{document}
\maketitle
\begin{abstract}

Contrastive explanations, which indicate why an AI system  produced one output (the target) instead of another (the foil), are widely regarded in explainable AI as more informative and interpretable than standard explanations. However, obtaining such explanations for speech-to-text (S2T) generative models remains an open challenge. Drawing from feature attribution techniques, we propose the first method to obtain contrastive explanations in S2T by analyzing how parts of the input spectrogram influence the choice between alternative outputs. Through a case study on gender assignment in speech translation, we show that our method accurately identifies the audio features that drive the selection of one gender over another.
By extending the scope of contrastive explanations to S2T, our work provides a foundation for better understanding S2T models.

\end{abstract}

\section{Introduction}

\begin{figure}[t]
    \centering
    \includegraphics[width=0.93\columnwidth]{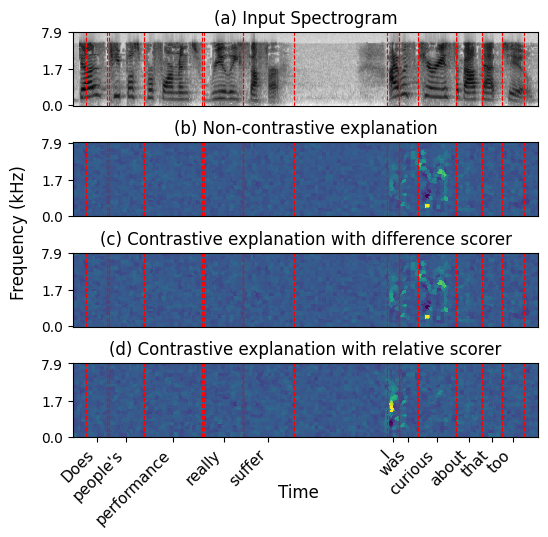}
    \caption{
    Input spectrogram (\textit{a}) for the utterance of ``Does people's performance really suffer? I was curious about that, too'' with three saliency maps (lighter = more relevant): a standard explanation for translating `\textit{curious}' as \textit{\underline{curiosa}} (b), and two contrastive explanations of why \textit{curios\textbf{\underline{a}}}$^F$ was preferred over \textit{curios\textbf{\underline{o}}}$^M$ obtained with the  difference scorer (c) or the relative scorer (d).
    }
    \label{fig:explanations}
\end{figure}

The rise of deep neural networks has increased the demand for explainable AI (XAI) methods to understand model behavior \cite{rauker2023toward, ferrando2024primer}. Within XAI, contrastive explanations---which aim to answer the question ‘\textit{Why did P happen rather than Q?}’ instead of simply ‘\textit{Why did P happen?}’ \cite{lipton1990contrastive}---have emerged as a promising approach with increasing adoption across various XAI applications \cite{stepin2021survey}. Their
advantage stems from mirroring human reasoning
\cite{byrne2002mental, miller2019explanation} and providing more targeted insights than traditional explanations \cite{lipton1990contrastive, jacovi-etal-2021-contrastive}.

Nevertheless, contrastive explanations have yet to be applied to speech-to-text (S2T) models. As S2T adoption grows \cite{latif2023sparks, barrault2025joint}, extending XAI advances like contrastive explanations to speech becomes crucial.
Some pioneering works have tackled the explanation of S2T models' decisions, braving key challenges including the multidimensional nature of speech signals, which span time and frequency, and the variable length of output sequences \cite{wu2024can}. The main approach used in the literature is to measure how perturbing the input audio signal affects the output \cite{mandel2016directly, kavaki2020identifying, trinh2020directly, markertvisualizing, mohebbi-etal-2023-homophone, wu2023explanations, fucci2024spes, wu2024can}. 
The resulting explanations take the form of saliency maps over a spectrogram representation of the audio input (Fig. \ref{fig:explanations}.a). These maps highlight the input regions that most strongly influence the model's predictions (Fig.~\ref{fig:explanations}.b). However, such explanations are holistic: they identify features relevant for all aspects of word generation (e.g.\ why the model produces `\underline{curioso}'), without focusing on specific contrastive aspects (why `curios\textbf{\underline{o}}', in the masculine, instead of `curios\textbf{\underline{a}}').
\textbf{How, then, can we obtain contrastive explanations of the use of speech features by S2T models?}

To meet this challenge, we build upon a prior non-contrastive feature attribution method for S2T, SPES \citep{fucci2024spes}. Adapting it to produce contrastive explanations requires two main interventions: \textit{i)} aggregating token-level probabilities for word-level analysis, since we aim to explain why one word was generated instead of another, while models generate subword tokens rather than complete words, and \textit{ii)} designing a scoring function that quantifies relative probability changes between a \textit{target} word and an alternative (the \textit{foil}). We investigate multiple approaches for both challenges, finding that standard solutions from text-based NLP are inadequate for our scenario, and propose improvements upon them.

We evaluate our method through a case study in speech translation (ST), focusing on the translation of gender-neutral terms referring to the speaker to languages requiring grammatical gender choices. For instance, translating ``I am curious'' to Italian typically involves choosing between an adjective with masculine inflection or one with feminine inflection. This setting is well suited for evaluating contrastive explanations in S2T as it provides natural pairs to contrast (masculine/feminine forms), while offering an opportunity to study whether models use acoustic cues like the speaker's pitch to disambiguate gender \cite{bentivogli-etal-2020-gender}.\footnote{We acknowledge that using vocal features like pitch for gender prediction and framing gender as a binary construct raise important ethical concerns, which we discuss in \S \ref{sec:ethics}.}

Our method produces markedly different saliency maps from non-contrastive approaches (see Fig. \ref{fig:explanations}.d), and our experiments confirm that these differences reflect its ability to isolate gender-specific features, while non-contrastive explanations highlight regions affecting general word prediction.
Our contributions are:
\textit{i)} \textbf{the first method for contrastive explanations in S2T} (\S\ref{scorers}), enabling precise identification of the input features a model relies on to translate gender-ambiguous terms referring to the speaker; \textit{ii)} \textbf{a methodology to evaluate their faithfulness} \S\ref{eval}); \textit{iii)} \textbf{empirical validation} on three language pairs (en$\rightarrow$fr/it/es) demonstrating that our method successfully isolates gender-relevant features (\S\ref{sec:use-case}).\footnote{Our code is available at https://github.com/hlt-mt/FBK-fairseq under the Apache License 2.0.}

\section{Contrastive Explanations for S2T}
\label{scorers}

We build on SPES \citep{fucci2024spes}, a state-of-the-art perturbation-based method for feature attribution in S2T models. SPES segments the input spectrogram by identifying acoustically meaningful regions based on intensity patterns. It then performs multiple inference passes, each time perturbing the input by randomly masking a subset of these segments by setting them to zero. The influence of each segment on the model's output is quantified by comparing the perturbed and original probabilities of a target token $t$, using a scoring function $S_B$.
The scores are averaged over all perturbations for each segment, resulting in a final saliency map that highlights the most relevant regions of the spectrogram for predicting $t$. These saliency maps explain individual token predictions, not word-level patterns, since S2T models generate subword tokens. 

Obtaining contrastive explanations, that identify why one word was chosen instead of another, requires two main changes to SPES:
\textit{i)} aggregating token-level probabilities for a word-level analysis (\S\ref{subsec:wlp}) and \textit{ii)} designing a scoring function to quantify relative changes in probability between target $t$ and foil $f$ (\S\ref{subsec:sf}).

\subsection{Word-Level Probabilities}
\label{subsec:wlp}

Thus, the probability of a word $t$ must be reconstructed from its $k$ constituent subwords $w_{0,\dots,k} \in t$. Previous work on feature attribution for text generation \cite{vafa-etal-2021-rationales, ferrando-etal-2022-towards, yin-neubig-2022-interpreting, sarti-etal-2023-inseq} relies on simplistic methods like subword-level attribution or basic aggregation techniques (averaging, max-pooling), failling to account for potential mismatches in the number of tokens in $t$ and $f$, and for the fact that a subword sequence may be the prefix of a longer word, rather a complete word.

\textbf{Our Solution.} To overcome the limitations of previous approaches and obtain accurate word-level probabilities, we adopt the methodology of \citet{pimentel-meister-2024-compute}.
Unlike other simplistic methods, this approach accounts for the fact that a sequence of subwords should be followed by a beginning of word or punctuation token to form a word rather than a prefix.
This distinction is crucial to avoid overestimating the likelihood of subword sequences that could be prefixes (e.g., Italian: \texttt{\_professor e}$^M$ vs. \texttt{\_professor e ssa}$^F$). 
To the best of our knowledge, we are the first to apply this principled method to feature attribution, ensuring accurate word-level explanations.

\subsection{Scoring Functions}
\label{subsec:sf}

In perturbation-based feature attribution, the base scoring function $S_{B}$ is generally defined as the difference between the original probability $p(t)$ and the probability $\tilde{p}(t)$ under the perturbed input \citep{ancona2018towards, seo2019regional}:
\begin{equation}
S_{B}(t) = p(t) - \tilde{p}(t)
\end{equation}
Instead, to explain why a model produced an output rather than another, contrastive feature attribution works in NLP \cite{eberle-etal-2023-rather, ferrando-etal-2023-explaining, sarti-etal-2023-inseq, sarti-etal-2023-quantifying, krishna2024post} typically use the contrastive \textit{difference scorer} by \citet{yin-neubig-2022-interpreting}:
\begin{equation}
    S_{CD}(t,f) = (p(t) - \tilde{p}(t)) - (p(f) - \tilde{p}(f))
\end{equation}
$S_{CD}(t,f)$ assigns high scores to features whose perturbation simultaneously decreases $p(t)$ and increases $p(f)$.
However, if the model strongly favors the target over the foil ($p(t) \gg p(f)$), the difference score $S_{CD}(t,f)$ becomes nearly identical to the base score $S_B(t)$.
As shown in Fig. \ref{fig:explanations}.c, this produces explanations indistinguishable from non-contrastive ones, highlighting features involved in every aspect of the prediction of $t$ rather than focusing on the contrastive aspect under study—a limitation also noted by \citet{eberle-etal-2023-rather} in NLP contexts.

\textbf{Our Solution.} To address this limitation, we repurpose the \textit{relative scorer} of \citet{jacovi-etal-2021-contrastive}, originally introduced to evaluate counterfactuals:
\begin{equation}
S_{CR}(t,f) = \frac{p(t)}{p(t) + p(f)} - \frac{\tilde{p}(t)}{\tilde{p}(t) + \tilde{p}(f)}
\end{equation}
This score normalizes the contribution of each term by their sum, which ensures that both terms remain influential in the final score even when their probabilities differ by orders of magnitude.
The resulting saliency maps (Fig. \ref{fig:explanations}.d) differ significantly from non-contrastive ones, indicating its precision in isolating gender-specific features. 
The theoretical advantages of $S_{CR}$'s probability ratio normalization extend beyond our S2T application, since generative models commonly produce cases where $p(t) \gg p(f)$.
Verifying whether $S_{CR}$ is also preferable to $S_{CD}$ for contrastive explanations of other generation tasks falls outside the scope of this work, but represents a promising direction for future research within the broader XAI community.

\section{Evaluation}\label{eval}

While existing work on contrastive explanations often evaluates against linguistically plausible human-annotated explanations \cite{yin-neubig-2022-interpreting, eberle-etal-2023-rather, ferrando-etal-2023-explaining}, we focus on faithfulness—how accurately explanations reflect model behavior, regardless of human interpretability \cite{jacovi-goldberg-2020-towards}.

We adapt the flip rate metric proposed by \citet{chemmengath-etal-2022-cat} for evaluating counterfactuals in text classification. In their work, the flip rate measures how often a counterfactual input causes the model to predict the foil instead of the target. Since our explanations are saliency maps rather than input variants, we combine this concept with the \textbf{deletion metric}—a common approach for evaluating feature attribution methods \cite{samek2016evaluating, arras2017explaining, tomsett2020sanity, samek2021explaining, fucci2024spes, gevaert2024evaluating}. Starting with the most salient regions identified by our method, we progressively remove input features and measure how quickly this causes the model to switch from predicting the target to the foil. A faithful explanation should identify features that require minimal deletion to flip the model's prediction.

However, unlike in classification tasks with fixed output classes, generative models can produce any text when perturbed, including synonyms or unrelated outputs. For our application, this complicates detecting when the predicted gender changes. Therefore, we separate our evaluation into two complementary metrics: \textbf{coverage}, which tracks the percentage of cases where the model generates either term of interest, and \textbf{flip rate}, which measures how often the model switches from target to foil within these covered cases. While the flip rate indicates whether we identified the right features for the target-foil contrast, coverage ensures these features are specific to the contrast rather than affecting general text generation.

\section{Use Case}
\label{sec:use-case}

We test our contrastive feature attribution method on speaker's gender assignment in S2T translation. 
As a \textbf{benchmark}, we use MuST-SHE \cite{bentivogli-etal-2020-gender}, which contains English-to-Spanish/French/Italian ST pairs annotated for terms that lack explicit gender marking in the source but require gender assignment in the target language.
We focus on terms referring to the speaker,\footnote{For details on example selection, see Appendix \ref{sec:data}.} for which MuST-SHE provides both correct gender translations (based on gold speaker gender labels) and their alternative incorrect versions\footnote{Ethical implications of this study are discussed in \S \ref{sec:ethics}.} (e.g., `curiosa$^F$' vs. `curioso$^M$') which we use to construct the target/foil pairs.
Here, we compare our relative scoring function with the difference scorer on the en$\rightarrow$fr split, using the multilingual S2T Transformer encoder-decoder model by \citet{wang-etal-2020-fairseq}, which has demonstrated strong performance in gender translation accuracy.
Experiments with different architectures and language pairs, showing the same trends, are reported in Appendix~\ref{sec:other-experiments}, while the word-level aggregation methods are compared in Appendix \ref{sec:word-probs}.

\begin{figure}[t]
\centering
  \includegraphics[width=0.8\columnwidth]{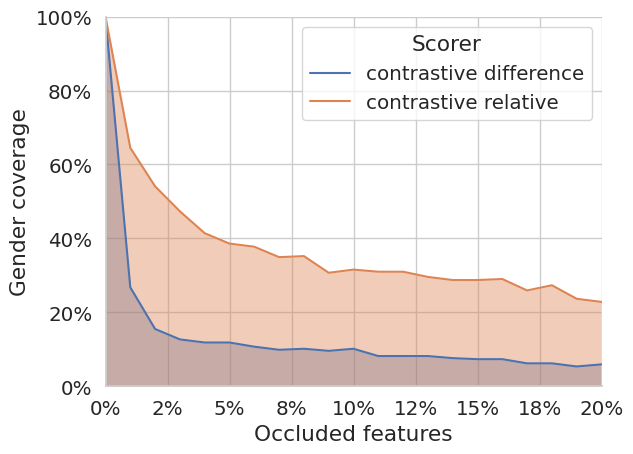}
  \caption{
  Coverage at different deletion steps for en$\rightarrow$fr.
  }
  \label{fig:transformer-fr-coverage}
\end{figure}

\paragraph{Coverage.} 
Fig. \ref{fig:transformer-fr-coverage} reports how coverage evolves as we progressively delete the features identified as most important.
The coverage of the difference scorer drops below 20\% after just 2\% deletion, while the relative scorer maintains higher coverage: more than 30\% even after 20\% deletion.\footnote{We focus on the first 20\% of feature deletion, as beyond this point the input becomes too degraded for meaningful model output. Full results are reported in Appendix \ref{sec:no-zoom}.}
The dramatic coverage loss of the difference scorer occurs because its explanations fail to be truly contrastive; they do not target gender-relevant features. 
This is demonstrated by their similarity to non-contrastive explanations:
saliency maps obtained with the contrastive difference scorer $S_{CD}$ on en-fr data exhibit a Pearson correlation coefficient of 0.93 with those obtained with the base scorer $S_B$, while relative scorer ($S_{CR}$) heatmaps show only 0.33 correlation (see Appendix \ref{sec:correlations} for other languages and models). 
We conclude that the difference scorer provides generic explanations, while \textbf{the features highlighted by the relative scorer are more precisely linked to gender assignment.}

\begin{figure}[t]
\centering
    \includegraphics[width=0.8\columnwidth]{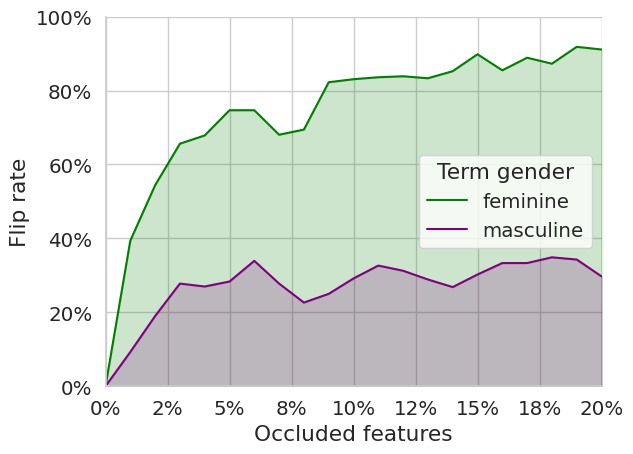}
    \caption{
    Flip rate at different deletion steps for en$\rightarrow$fr.}
    \label{fig:transformer-fr-swap}
\end{figure}

\paragraph{Flip rate.}
\label{sec:results_flip}

We now demonstrate that the features identified by the relative scorer are responsible for shifting the model's predictions in the expected direction along the gender axis.
Fig. \ref{fig:transformer-fr-swap} shows that the relative scorer effectively isolates features driving feminine gender predictions: occluding the top 5\% most relevant features causes the model to switch to masculine translations for over 70\% of the terms that remain in coverage. This high flip rate, combined with the maintained coverage discussed above, demonstrates that \textbf{our method  precisely discriminates  the  exact input regions driving translation toward one gender instead of the other}: the model generates the same terms (coverage) but changes their gender (flip rate) when these features are removed. 
For masculine predictions, in contrast, the flip rate is much lower, plateauing at $\sim$30\%.
However, this may not indicate a limitation of the method, but rather reflect an underlying property of the ST model: if masculine serves as the default prediction rather than one actively triggered by specific input cues, then no occluded features could shift the prediction toward the feminine class.
The XAI literature on gender bias supports this hypothesis: \citet{jumelet-etal-2019-analysing} observe that models use masculine as a default, generating feminine forms only when strong feminine signals are present. This masculine default bias has also been observed in ST systems' internal language models \citep{fucci-etal-2023-integrating} and likely stems from training data gender imbalance \cite{tatman-2017-gender}---MuST-C contains twice as many male as female speakers \cite{cattoni2021must}.
We defer to future work a thorough analysis of the acoustic features the model exploits for gender translation and their interaction with other model components and potential sources of bias.

Overall, our method effectively isolates features involved in gender prediction, disentangling them from those encoding other semantic content.

\section{Conclusion}

This paper has introduced the first method to obtain contrastive explanations for S2T models, which identify input features that lead the model to generate one output word instead of another. Our solution builds on perturbation-based feature attribution over spectrograms. The key element is our scorer, which ensures truly contrastive explanations, unlike the difference scorer widely used in NLP. Our case study on the input features driving the translation of speakers' gender in ST demonstrated the effectiveness of our approach in isolating the input features driving the model’s choice between feminine and masculine grammatical forms. Our methodology will enable future research not only to investigate which phonetic cues ST models use for gender disambiguation, but could also be applied to other phenomena and S2T tasks.

\section{Limitations}

\paragraph{Extension to other use cases}
While this work demonstrates the effectiveness of the proposed contrastive explanation methodology for analyzing gender translation, the method's applicability beyond gender analysis remains to be validated across other phenomena. However, we do not foresee any reason for which its applicability could be limited.
Possible areas of investigation include:
i) understanding homophone disambiguation (e.g., ``plain'' vs. ``plane'') based on audio features \cite{mohebbi-etal-2023-homophone, yu-etal-2024-speech}, ii) analyzing how models resolve coreference resolution (e.g., identifying the features that make a model associate an ambiguous pronoun with one referent over another) \cite{amoia2012coreference, roesiger-etal-2017-improving}, iii) investigating factuality issues by contrasting correct translations against model errors, and iv) exploring politeness register selection (e.g., why models choose formal ``vous/Sie'' over informal ``tu/du'') \cite{sennrich-etal-2016-controlling}.

\paragraph{Explaining gender assignment}
As discussed 
\df{in \S \ref{sec:results_flip},} 
the flip rate for feminine and masculine forms is different, and never reaches 100\%. For the Conformer models in Appendix \ref{sec:other-experiments}, the flip rate is closer among genders.
We propose potential reasons for these variations, including differences in models' initial gender translation capabilities and possible masculine default bias in ST models' internal language models. But these remain conjectures that require empirical verification. Such investigation of how different ST models translate gender falls outside the scope of this work, which focuses on developing and evaluating contrastive explanations rather than reaching conclusions about gender translation mechanisms in ST models.

\section{Ethics Statement}
\label{sec:ethics}

\df{The \textbf{reliance of ST systems on vocal traits for gender prediction}, as examined in our use case, raises ethical concerns. These vocal differences arise from both physiological and sociocultural factors, which are well-documented in the literature (e.g., \citealp{simpson_2001,hillenbrand-2009-pitch,coleman-1976-voice,nylen_2004}). While leveraging such traits has been shown to improve overall gender translation accuracy in ST \cite{bentivogli-etal-2020-gender}, it may also disadvantage certain minorities and vulnerable populations.}
Transgender speakers and individuals with vocal impairments could receive lower quality translations if their vocal characteristics do not align with typical gender-based expectations \cite{matar2016voice, pereira2018percepccao, villas2021acoustic, menezes2024prosodic}. Prescribing how ST systems should handle gender-related translations or which characteristics they should consider when making gender-related decisions is beyond the scope of this paper. Instead, our work focuses on explaining current model behaviors and understanding how these systems operate when left to make their own decisions.

Second, our contrastive study is constrained by \textbf{the binary (masculine/feminine) gender} annotations in our dataset \cite{bentivogli-etal-2020-gender}. While this binary framework reflects the grammatical conventions of our target languages, from a social perspective, it fails to acknowledge individuals whose gender identity exists outside the male/female binary \cite{10.1093/oxfordhb/9780190212926.013.45}. Although we analyze grammatical gender, we acknowledge its interaction with gender identity and recognize that enforcing binary translations can erase non-binary identities. While emerging linguistic innovations like neologisms, neopronouns, and new morphological forms offer alternatives \cite{piergentili-etal-2023-gender}, to the best of our knowledge, ST datasets incorporating these forms are not yet available. Once such data becomes available, our methodology could be used to analyze how systems choose between binary terms and neutral alternatives. Furthermore, extending this analysis to languages with diverse gender systems could provide valuable insights into how ST models handle varying degrees of grammatical gender complexity \cite{corbett1991gender}.

\section*{Acknowledgements}

This paper has received funding from the PNRR project FAIR - Future AI Research (PE00000013),  under the NRRP MUR program funded by the NextGenerationEU and from the European Union’s Horizon research and innovation programme under grant agreement No 101135798, project Meetween (My Personal AI Mediator for Virtual MEETings BetWEEN People).

\bibliography{anthology,custom}

\begin{thebibliography}{69}
\providecommand{\natexlab}[1]{#1}

\bibitem[{Achanta et~al.(2012)Achanta, Shaji, Smith, Lucchi, Fua, and S{\"u}sstrunk}]{achanta2012slic}
Radhakrishna Achanta, Appu Shaji, Kevin Smith, Aurelien Lucchi, Pascal Fua, and Sabine S{\"u}sstrunk. 2012.
\newblock Slic superpixels compared to state-of-the-art superpixel methods.
\newblock \emph{IEEE transactions on pattern analysis and machine intelligence}, 34(11):2274--2282.

\bibitem[{Ahmad et~al.(2024)Ahmad, Anastasopoulos, Bojar, Borg, Carpuat, Cattoni, Cettolo, Chen, Dong, Federico, Haddow, Javorsk{\'y}, Krubi{\'n}ski, Lam, Ma, Mathur, Matusov, Maurya, McCrae, Murray, Nakamura, Negri, Niehues, Niu, Ojha, Ortega, Papi, Pol{\'a}k, Posp{\'i}{\v{s}}il, Pecina, Salesky, Sethiya, Sarkar, Shi, Sikasote, Sperber, St{\"u}ker, Sudoh, Thompson, Waibel, Watanabe, Wilken, Zem{\'a}nek, and Zevallos}]{ahmad-etal-2024-findings}
Ibrahim~Said Ahmad, Antonios Anastasopoulos, Ond{\v{r}}ej Bojar, Claudia Borg, Marine Carpuat, Roldano Cattoni, Mauro Cettolo, William Chen, Qianqian Dong, Marcello Federico, Barry Haddow, D{\'a}vid Javorsk{\'y}, Mateusz Krubi{\'n}ski, Tsz~Kin Lam, Xutai Ma, Prashant Mathur, Evgeny Matusov, Chandresh Maurya, John McCrae, Kenton Murray, Satoshi Nakamura, Matteo Negri, Jan Niehues, Xing Niu, Atul~Kr. Ojha, John Ortega, Sara Papi, Peter Pol{\'a}k, Adam Posp{\'i}{\v{s}}il, Pavel Pecina, Elizabeth Salesky, Nivedita Sethiya, Balaram Sarkar, Jiatong Shi, Claytone Sikasote, Matthias Sperber, Sebastian St{\"u}ker, Katsuhito Sudoh, Brian Thompson, Alex Waibel, Shinji Watanabe, Patrick Wilken, Petr Zem{\'a}nek, and Rodolfo Zevallos. 2024.
\newblock \href {https://doi.org/10.18653/v1/2024.iwslt-1.1} {{FINDINGS} {OF} {THE} {IWSLT} 2024 {EVALUATION} {CAMPAIGN}}.
\newblock In \emph{Proceedings of the 21st International Conference on Spoken Language Translation (IWSLT 2024)}, pages 1--11, Bangkok, Thailand (in-person and online). Association for Computational Linguistics.

\bibitem[{Amoia et~al.(2012)Amoia, Kunz, and Lapshinova-Koltunski}]{amoia2012coreference}
Marilisa Amoia, Kerstin Kunz, and Ekaterina Lapshinova-Koltunski. 2012.
\newblock Coreference in spoken vs. written texts: a corpus-based analysis.
\newblock In \emph{LREC}, pages 158--164.

\bibitem[{Ancona et~al.(2018)Ancona, Ceolini, {\"O}ztireli, and Gross}]{ancona2018towards}
Marco Ancona, Enea Ceolini, Cengiz {\"O}ztireli, and Markus Gross. 2018.
\newblock Towards better understanding of gradient-based attribution methods for deep neural networks.
\newblock In \emph{6th International Conference on Learning Representations (ICLR)}, 1711.06104, pages 0--0. Arxiv-Computer Science.

\bibitem[{Ardila et~al.(2020)Ardila, Branson, Davis, Kohler, Meyer, Henretty, Morais, Saunders, Tyers, and Weber}]{ardila-etal-2020-common}
Rosana Ardila, Megan Branson, Kelly Davis, Michael Kohler, Josh Meyer, Michael Henretty, Reuben Morais, Lindsay Saunders, Francis Tyers, and Gregor Weber. 2020.
\newblock \href {https://aclanthology.org/2020.lrec-1.520/} {Common voice: A massively-multilingual speech corpus}.
\newblock In \emph{Proceedings of the Twelfth Language Resources and Evaluation Conference}, pages 4218--4222, Marseille, France. European Language Resources Association.

\bibitem[{Arras et~al.(2017)Arras, Montavon, M{\"u}ller, and Samek}]{arras2017explaining}
Leila Arras, Gr{\'e}goire Montavon, Klaus-Robert M{\"u}ller, and Wojciech Samek. 2017.
\newblock Explaining recurrent neural network predictions in sentiment analysis.
\newblock In \emph{8th Workshop on Computational Approaches to Subjectivity, Sentiment and Social Media Analysis WASSA 2017: Proceedings of the Workshop}, pages 159--168. The Association for Computational Linguistics.

\bibitem[{Barrault et~al.(2025)Barrault, Chung, Meglioli, Dale, Dong, Duquenne, Elsahar, Gong, Heffernan, Hoffman et~al.}]{barrault2025joint}
Lo{\"\i}c Barrault, Yu-An Chung, Mariano~Coria Meglioli, David Dale, Ning Dong, Paul-Ambroise Duquenne, Hady Elsahar, Hongyu Gong, Kevin Heffernan, John Hoffman, et~al. 2025.
\newblock Joint speech and text machine translation for up to 100 languages.
\newblock \emph{Nature}, 637(8046):587--593.

\bibitem[{Bentivogli et~al.(2020)Bentivogli, Savoldi, Negri, Di~Gangi, Cattoni, and Turchi}]{bentivogli-etal-2020-gender}
Luisa Bentivogli, Beatrice Savoldi, Matteo Negri, Mattia~A. Di~Gangi, Roldano Cattoni, and Marco Turchi. 2020.
\newblock \href {https://doi.org/10.18653/v1/2020.acl-main.619} {Gender in danger? evaluating speech translation technology on the {M}u{ST}-{SHE} corpus}.
\newblock In \emph{Proceedings of the 58th Annual Meeting of the Association for Computational Linguistics}, pages 6923--6933, Online. Association for Computational Linguistics.

\bibitem[{Byrne(2002)}]{byrne2002mental}
Ruth~MJ Byrne. 2002.
\newblock Mental models and counterfactual thoughts about what might have been.
\newblock \emph{Trends in cognitive sciences}, 6(10):426--431.

\bibitem[{Cattoni et~al.(2021)Cattoni, Di~Gangi, Bentivogli, Negri, and Turchi}]{cattoni2021must}
Roldano Cattoni, Mattia~Antonino Di~Gangi, Luisa Bentivogli, Matteo Negri, and Marco Turchi. 2021.
\newblock Must-c: A multilingual corpus for end-to-end speech translation.
\newblock \emph{Computer speech \& language}, 66:101155.

\bibitem[{Chemmengath et~al.(2022)Chemmengath, Azad, Luss, and Dhurandhar}]{chemmengath-etal-2022-cat}
Saneem Chemmengath, Amar~Prakash Azad, Ronny Luss, and Amit Dhurandhar. 2022.
\newblock \href {https://doi.org/10.18653/v1/2022.emnlp-main.484} {Let the {CAT} out of the bag: Contrastive attributed explanations for text}.
\newblock In \emph{Proceedings of the 2022 Conference on Empirical Methods in Natural Language Processing}, pages 7190--7206, Abu Dhabi, United Arab Emirates. Association for Computational Linguistics.

\bibitem[{Coleman(1976)}]{coleman-1976-voice}
Ralph~O. Coleman. 1976.
\newblock \href {https://doi.org/10.1044/jshr.1901.168} {A comparison of the contributions of two voice quality characteristics to the perception of maleness and femaleness in the voice}.
\newblock \emph{Journal of Speech \& Hearing Research}, 19(1):168--180.

\bibitem[{Corbett(1991)}]{corbett1991gender}
Greville~G Corbett. 1991.
\newblock \emph{Gender}.
\newblock Cambridge University Press.

\bibitem[{Eberle et~al.(2023)Eberle, Chalkidis, Cabello, and Brandl}]{eberle-etal-2023-rather}
Oliver Eberle, Ilias Chalkidis, Laura Cabello, and Stephanie Brandl. 2023.
\newblock \href {https://doi.org/10.18653/v1/2023.emnlp-main.427} {Rather a nurse than a physician - contrastive explanations under investigation}.
\newblock In \emph{Proceedings of the 2023 Conference on Empirical Methods in Natural Language Processing}, pages 6907--6920, Singapore. Association for Computational Linguistics.

\bibitem[{Ferrando et~al.(2022)Ferrando, G{\'a}llego, Alastruey, Escolano, and Costa-juss{\`a}}]{ferrando-etal-2022-towards}
Javier Ferrando, Gerard~I. G{\'a}llego, Belen Alastruey, Carlos Escolano, and Marta~R. Costa-juss{\`a}. 2022.
\newblock \href {https://doi.org/10.18653/v1/2022.emnlp-main.599} {Towards opening the black box of neural machine translation: Source and target interpretations of the transformer}.
\newblock In \emph{Proceedings of the 2022 Conference on Empirical Methods in Natural Language Processing}, pages 8756--8769, Abu Dhabi, United Arab Emirates. Association for Computational Linguistics.

\bibitem[{Ferrando et~al.(2023)Ferrando, G{\'a}llego, Tsiamas, and Costa-juss{\`a}}]{ferrando-etal-2023-explaining}
Javier Ferrando, Gerard~I. G{\'a}llego, Ioannis Tsiamas, and Marta~R. Costa-juss{\`a}. 2023.
\newblock \href {https://doi.org/10.18653/v1/2023.acl-long.301} {Explaining how transformers use context to build predictions}.
\newblock In \emph{Proceedings of the 61st Annual Meeting of the Association for Computational Linguistics (Volume 1: Long Papers)}, pages 5486--5513, Toronto, Canada. Association for Computational Linguistics.

\bibitem[{Ferrando et~al.(2024)Ferrando, Sarti, Bisazza, and Costa-juss{\`a}}]{ferrando2024primer}
Javier Ferrando, Gabriele Sarti, Arianna Bisazza, and Marta~R Costa-juss{\`a}. 2024.
\newblock A primer on the inner workings of transformer-based language models.
\newblock \emph{arXiv preprint arXiv:2405.00208}.

\bibitem[{Fucci et~al.(2023)Fucci, Gaido, Papi, Cettolo, Negri, and Bentivogli}]{fucci-etal-2023-integrating}
Dennis Fucci, Marco Gaido, Sara Papi, Mauro Cettolo, Matteo Negri, and Luisa Bentivogli. 2023.
\newblock \href {https://doi.org/10.18653/v1/2023.emnlp-main.705} {Integrating language models into direct speech translation: An inference-time solution to control gender inflection}.
\newblock In \emph{Proceedings of the 2023 Conference on Empirical Methods in Natural Language Processing}, pages 11505--11517, Singapore. Association for Computational Linguistics.

\bibitem[{Fucci et~al.(2024)Fucci, Gaido, Savoldi, Negri, Cettolo, and Bentivogli}]{fucci2024spes}
Dennis Fucci, Marco Gaido, Beatrice Savoldi, Matteo Negri, Mauro Cettolo, and Luisa Bentivogli. 2024.
\newblock Spes: Spectrogram perturbation for explainable speech-to-text generation.
\newblock \emph{arXiv preprint arXiv:2411.01710}.

\bibitem[{Gevaert et~al.(2024)Gevaert, Rousseau, Becker, Valkenborg, De~Bie, and Saeys}]{gevaert2024evaluating}
Arne Gevaert, Axel-Jan Rousseau, Thijs Becker, Dirk Valkenborg, Tijl De~Bie, and Yvan Saeys. 2024.
\newblock Evaluating feature attribution methods in the image domain.
\newblock \emph{Machine Learning}, pages 1--46.

\bibitem[{Hernandez et~al.(2018)Hernandez, Nguyen, Ghannay, Tomashenko, and Est{\`e}ve}]{tedlium}
Fran{\c{c}}ois Hernandez, Vincent Nguyen, Sahar Ghannay, Natalia Tomashenko, and Yannick Est{\`e}ve. 2018.
\newblock Ted-lium 3: Twice as much data and corpus repartition for experiments on speaker adaptation.
\newblock In \emph{Speech and Computer}, pages 198--208, Cham.

\bibitem[{Hillenbrand and Clark(2009)}]{hillenbrand-2009-pitch}
James~M. Hillenbrand and Michael~J. Clark. 2009.
\newblock \href {https://doi.org/10.3758/APP.71.5.1150} {The role of f0 and formant frequencies in distinguishing the voices of men and women}.
\newblock \emph{Attention Perception \& Psychophysics}, 71(5):1150--1166.

\bibitem[{Iranzo-Sánchez et~al.(2020)Iranzo-Sánchez, Silvestre-Cerdà, Jorge, Roselló, Giménez, Sanchis, Civera, and Juan}]{europarlst}
Javier Iranzo-Sánchez, Joan~Albert Silvestre-Cerdà, Javier Jorge, Nahuel Roselló, Adrià Giménez, Albert Sanchis, Jorge Civera, and Alfons Juan. 2020.
\newblock \href {https://doi.org/10.1109/ICASSP40776.2020.9054626} {Europarl-st: A multilingual corpus for speech translation of parliamentary debates}.
\newblock In \emph{ICASSP 2020 - 2020 IEEE International Conference on Acoustics, Speech and Signal Processing (ICASSP)}, pages 8229--8233.

\bibitem[{Jacovi and Goldberg(2020)}]{jacovi-goldberg-2020-towards}
Alon Jacovi and Yoav Goldberg. 2020.
\newblock \href {https://doi.org/10.18653/v1/2020.acl-main.386} {Towards faithfully interpretable {NLP} systems: How should we define and evaluate faithfulness?}
\newblock In \emph{Proceedings of the 58th Annual Meeting of the Association for Computational Linguistics}, pages 4198--4205, Online. Association for Computational Linguistics.

\bibitem[{Jacovi et~al.(2021)Jacovi, Swayamdipta, Ravfogel, Elazar, Choi, and Goldberg}]{jacovi-etal-2021-contrastive}
Alon Jacovi, Swabha Swayamdipta, Shauli Ravfogel, Yanai Elazar, Yejin Choi, and Yoav Goldberg. 2021.
\newblock \href {https://doi.org/10.18653/v1/2021.emnlp-main.120} {Contrastive explanations for model interpretability}.
\newblock In \emph{Proceedings of the 2021 Conference on Empirical Methods in Natural Language Processing}, pages 1597--1611, Online and Punta Cana, Dominican Republic. Association for Computational Linguistics.

\bibitem[{Jumelet et~al.(2019)Jumelet, Zuidema, and Hupkes}]{jumelet-etal-2019-analysing}
Jaap Jumelet, Willem Zuidema, and Dieuwke Hupkes. 2019.
\newblock \href {https://doi.org/10.18653/v1/K19-1001} {Analysing neural language models: Contextual decomposition reveals default reasoning in number and gender assignment}.
\newblock In \emph{Proceedings of the 23rd Conference on Computational Natural Language Learning (CoNLL)}, pages 1--11, Hong Kong, China. Association for Computational Linguistics.

\bibitem[{Kavaki and Mandel(2020)}]{kavaki2020identifying}
Hassan~Salami Kavaki and Michael~I Mandel. 2020.
\newblock Identifying important time-frequency locations in continuous speech utterances.
\newblock In \emph{Proceedings of Interspeech}.

\bibitem[{Krishna et~al.(2024)Krishna, Ma, Slack, Ghandeharioun, Singh, and Lakkaraju}]{krishna2024post}
Satyapriya Krishna, Jiaqi Ma, Dylan Slack, Asma Ghandeharioun, Sameer Singh, and Himabindu Lakkaraju. 2024.
\newblock Post hoc explanations of language models can improve language models.
\newblock \emph{Advances in Neural Information Processing Systems}, 36.

\bibitem[{Latif et~al.(2023)Latif, Shoukat, Shamshad, Usama, Ren, Cuay{\'a}huitl, Wang, Zhang, Togneri, Cambria et~al.}]{latif2023sparks}
Siddique Latif, Moazzam Shoukat, Fahad Shamshad, Muhammad Usama, Yi~Ren, Heriberto Cuay{\'a}huitl, Wenwu Wang, Xulong Zhang, Roberto Togneri, Erik Cambria, et~al. 2023.
\newblock Sparks of large audio models: A survey and outlook.
\newblock \emph{arXiv preprint arXiv:2308.12792}.

\bibitem[{Lipton(1990)}]{lipton1990contrastive}
Peter Lipton. 1990.
\newblock Contrastive explanation.
\newblock \emph{Royal Institute of Philosophy Supplements}, 27:247--266.

\bibitem[{Mandel(2016)}]{mandel2016directly}
Michael~I Mandel. 2016.
\newblock Directly comparing the listening strategies of humans and machines.
\newblock In \emph{INTERSPEECH}, pages 660--664.

\bibitem[{Markert et~al.(2021)Markert, Parracone, Kulakov, Sperl, Kao, and B{\"o}ttinger}]{markertvisualizing}
Karla Markert, Romain Parracone, Mykhailo Kulakov, Philip Sperl, Ching-Yu Kao, and Konstantin B{\"o}ttinger. 2021.
\newblock Visualizing automatic speech recognition--means for a better understanding?
\newblock \emph{ISCA Symposium on Security and Privacy in Speech Communication}.

\bibitem[{Matar et~al.(2016)Matar, Portes, Lancia, Legou, and Baider}]{matar2016voice}
Nayla Matar, Cristel Portes, Leonardo Lancia, Thierry Legou, and Fabienne Baider. 2016.
\newblock Voice quality and gender stereotypes: A study of lebanese women with reinke's edema.
\newblock \emph{Journal of Speech, Language, and Hearing Research}, 59(6):S1608--S1617.

\bibitem[{Meister et~al.(2020)Meister, Cotterell, and Vieira}]{meister-etal-2020-beam}
Clara Meister, Ryan Cotterell, and Tim Vieira. 2020.
\newblock \href {https://doi.org/10.18653/v1/2020.emnlp-main.170} {If beam search is the answer, what was the question?}
\newblock In \emph{Proceedings of the 2020 Conference on Empirical Methods in Natural Language Processing (EMNLP)}, pages 2173--2185, Online. Association for Computational Linguistics.

\bibitem[{Menezes et~al.(2024)Menezes, de~Lira, de~Ara{\'u}jo, de~Almeida, Gomes, Moraes, and Lucena}]{menezes2024prosodic}
Danielle~Pereira Menezes, Zulina~Souza de~Lira, Ana Nery~Barbosa de~Ara{\'u}jo, Anna Alice~Figueir{\^e}do de~Almeida, Adriana de Oliveira~Camargo Gomes, Bruno~Teixeira Moraes, and Jonia~Alves Lucena. 2024.
\newblock Prosodic differences in the voices of transgender and cisgender women: self-perception of voice-an auditory and acoustic analysis.
\newblock \emph{Journal of Voice}, 38(4):844--857.

\bibitem[{Miller(2019)}]{miller2019explanation}
Tim Miller. 2019.
\newblock Explanation in artificial intelligence: Insights from the social sciences.
\newblock \emph{Artificial intelligence}, 267:1--38.

\bibitem[{Mohebbi et~al.(2023)Mohebbi, Chrupa{\l}a, Zuidema, and Alishahi}]{mohebbi-etal-2023-homophone}
Hosein Mohebbi, Grzegorz Chrupa{\l}a, Willem Zuidema, and Afra Alishahi. 2023.
\newblock \href {https://doi.org/10.18653/v1/2023.emnlp-main.513} {Homophone disambiguation reveals patterns of context mixing in speech transformers}.
\newblock In \emph{Proceedings of the 2023 Conference on Empirical Methods in Natural Language Processing}, pages 8249--8260, Singapore. Association for Computational Linguistics.

\bibitem[{Nylén et~al.(2024)Nylén, Holmberg, and Södersten}]{nylen_2004}
Frida Nylén, Johan Holmberg, and Maria Södersten. 2024.
\newblock \href {https://doi.org/10.1121/10.0025932} {Acoustic cues to femininity and masculinity in spontaneous speech}.
\newblock \emph{The Journal of the Acoustical Society of America}, 155(5):3090--3100.

\bibitem[{Panayotov et~al.(2015)Panayotov, Chen, Povey, and Khudanpur}]{librispeech}
Vassil Panayotov, Guoguo Chen, Daniel Povey, and Sanjeev Khudanpur. 2015.
\newblock \href {https://doi.org/10.1109/ICASSP.2015.7178964} {Librispeech: An asr corpus based on public domain audio books}.
\newblock In \emph{2015 IEEE International Conference on Acoustics, Speech and Signal Processing (ICASSP)}, pages 5206--5210.

\bibitem[{Papi et~al.(2024)Papi, Gaido, Pilzer, and Negri}]{papi-etal-2024-good}
Sara Papi, Marco Gaido, Andrea Pilzer, and Matteo Negri. 2024.
\newblock \href {https://doi.org/10.18653/v1/2024.acl-long.200} {When good and reproducible results are a giant with feet of clay: The importance of software quality in {NLP}}.
\newblock In \emph{Proceedings of the 62nd Annual Meeting of the Association for Computational Linguistics (Volume 1: Long Papers)}, pages 3657--3672, Bangkok, Thailand. Association for Computational Linguistics.

\bibitem[{Park et~al.(2019)Park, Chan, Zhang, Chiu, Zoph, Cubuk, and Le}]{park2019specaugment}
Daniel~S Park, William Chan, Yu~Zhang, Chung-Cheng Chiu, Barret Zoph, Ekin~D Cubuk, and Quoc~V Le. 2019.
\newblock Specaugment: A simple data augmentation method for automatic speech recognition.
\newblock \emph{Interspeech 2019}, page 2613.

\bibitem[{Pereira et~al.(2018)Pereira, Dassie-Leite, Pereira, Cavichiolo, Rosa, and Fugmann}]{pereira2018percepccao}
Amanda~Maria Pereira, Ana~Paula Dassie-Leite, Eliane~Cristina Pereira, Juliana~Benthien Cavichiolo, Marcelo de~Oliveira Rosa, and Elmar~Allen Fugmann. 2018.
\newblock Percep{\c{c}}{\~a}o auditiva de ju{\'\i}zes leigos quanto ao g{\^e}nero de mulheres com edema de reinke.
\newblock In \emph{CoDAS}, volume~30, page e20170046. SciELO Brasil.

\bibitem[{Piergentili et~al.(2023)Piergentili, Fucci, Savoldi, Bentivogli, and Negri}]{piergentili-etal-2023-gender}
Andrea Piergentili, Dennis Fucci, Beatrice Savoldi, Luisa Bentivogli, and Matteo Negri. 2023.
\newblock \href {https://aclanthology.org/2023.gitt-1.7/} {Gender neutralization for an inclusive machine translation: from theoretical foundations to open challenges}.
\newblock In \emph{Proceedings of the First Workshop on Gender-Inclusive Translation Technologies}, pages 71--83, Tampere, Finland. European Association for Machine Translation.

\bibitem[{Pimentel and Meister(2024)}]{pimentel-meister-2024-compute}
Tiago Pimentel and Clara Meister. 2024.
\newblock \href {https://doi.org/10.18653/v1/2024.emnlp-main.1020} {How to compute the probability of a word}.
\newblock In \emph{Proceedings of the 2024 Conference on Empirical Methods in Natural Language Processing}, pages 18358--18375, Miami, Florida, USA. Association for Computational Linguistics.

\bibitem[{R{\"a}uker et~al.(2023)R{\"a}uker, Ho, Casper, and Hadfield-Menell}]{rauker2023toward}
Tilman R{\"a}uker, Anson Ho, Stephen Casper, and Dylan Hadfield-Menell. 2023.
\newblock Toward transparent ai: A survey on interpreting the inner structures of deep neural networks.
\newblock In \emph{2023 ieee conference on secure and trustworthy machine learning (satml)}, pages 464--483. IEEE.

\bibitem[{Roesiger et~al.(2017)Roesiger, Stehwien, Riester, and Vu}]{roesiger-etal-2017-improving}
Ina Roesiger, Sabrina Stehwien, Arndt Riester, and Ngoc~Thang Vu. 2017.
\newblock \href {https://doi.org/10.18653/v1/W17-4610} {Improving coreference resolution with automatically predicted prosodic information}.
\newblock In \emph{Proceedings of the Workshop on Speech-Centric Natural Language Processing}, pages 78--83, Copenhagen, Denmark. Association for Computational Linguistics.

\bibitem[{Samek et~al.(2016)Samek, Binder, Montavon, Lapuschkin, and M{\"u}ller}]{samek2016evaluating}
Wojciech Samek, Alexander Binder, Gr{\'e}goire Montavon, Sebastian Lapuschkin, and Klaus-Robert M{\"u}ller. 2016.
\newblock Evaluating the visualization of what a deep neural network has learned.
\newblock \emph{IEEE transactions on neural networks and learning systems}, 28(11):2660--2673.

\bibitem[{Samek et~al.(2021)Samek, Montavon, Lapuschkin, Anders, and M{\"u}ller}]{samek2021explaining}
Wojciech Samek, Gr{\'e}goire Montavon, Sebastian Lapuschkin, Christopher~J Anders, and Klaus-Robert M{\"u}ller. 2021.
\newblock Explaining deep neural networks and beyond: A review of methods and applications.
\newblock \emph{Proceedings of the IEEE}, 109(3):247--278.

\bibitem[{Sarti et~al.(2024)Sarti, Chrupa{\l}a, Nissim, and Bisazza}]{sarti-etal-2023-quantifying}
Gabriele Sarti, Grzegorz Chrupa{\l}a, Malvina Nissim, and Arianna Bisazza. 2024.
\newblock \href {https://openreview.net/forum?id=XTHfNGI3zT} {Quantifying the plausibility of context reliance in neural machine translation}.
\newblock In \emph{The Twelfth International Conference on Learning Representations (ICLR 2024)}, Vienna, Austria. OpenReview.

\bibitem[{Sarti et~al.(2023)Sarti, Feldhus, Sickert, and van~der Wal}]{sarti-etal-2023-inseq}
Gabriele Sarti, Nils Feldhus, Ludwig Sickert, and Oskar van~der Wal. 2023.
\newblock \href {https://doi.org/10.18653/v1/2023.acl-demo.40} {Inseq: An interpretability toolkit for sequence generation models}.
\newblock In \emph{Proceedings of the 61st Annual Meeting of the Association for Computational Linguistics (Volume 3: System Demonstrations)}, pages 421--435, Toronto, Canada. Association for Computational Linguistics.

\bibitem[{Savoldi et~al.(2022)Savoldi, Gaido, Bentivogli, Negri, and Turchi}]{savoldi-etal-2022-morphosyntactic}
Beatrice Savoldi, Marco Gaido, Luisa Bentivogli, Matteo Negri, and Marco Turchi. 2022.
\newblock \href {https://doi.org/10.18653/v1/2022.acl-long.127} {Under the morphosyntactic lens: A multifaceted evaluation of gender bias in speech translation}.
\newblock In \emph{Proceedings of the 60th Annual Meeting of the Association for Computational Linguistics (Volume 1: Long Papers)}, pages 1807--1824, Dublin, Ireland. Association for Computational Linguistics.

\bibitem[{Sennrich et~al.(2016)Sennrich, Haddow, and Birch}]{sennrich-etal-2016-controlling}
Rico Sennrich, Barry Haddow, and Alexandra Birch. 2016.
\newblock \href {https://doi.org/10.18653/v1/N16-1005} {Controlling politeness in neural machine translation via side constraints}.
\newblock In \emph{Proceedings of the 2016 Conference of the North {A}merican Chapter of the Association for Computational Linguistics: Human Language Technologies}, pages 35--40, San Diego, California. Association for Computational Linguistics.

\bibitem[{Seo et~al.(2019)Seo, Oh, and Oh}]{seo2019regional}
Dasom Seo, Kanghan Oh, and Il-Seok Oh. 2019.
\newblock Regional multi-scale approach for visually pleasing explanations of deep neural networks.
\newblock \emph{IEEE Access}, 8:8572--8582.

\bibitem[{Simpson(2001)}]{simpson_2001}
Adrian~P. Simpson. 2001.
\newblock \href {https://doi.org/10.1121/1.1356020} {Dynamic consequences of differences in male and female vocal tract dimensions}.
\newblock \emph{The Journal of the Acoustical Society of America}, 109(5 Pt 1):2153--2164.

\bibitem[{Stepin et~al.(2021)Stepin, Alonso, Catala, and Pereira-Fari{\~n}a}]{stepin2021survey}
Ilia Stepin, Jose~M Alonso, Alejandro Catala, and Mart{\'\i}n Pereira-Fari{\~n}a. 2021.
\newblock A survey of contrastive and counterfactual explanation generation methods for explainable artificial intelligence.
\newblock \emph{IEEE Access}, 9:11974--12001.

\bibitem[{Tatman(2017)}]{tatman-2017-gender}
Rachael Tatman. 2017.
\newblock \href {https://doi.org/10.18653/v1/W17-1606} {Gender and dialect bias in {Y}ou{T}ube`s automatic captions}.
\newblock In \emph{Proceedings of the First {ACL} Workshop on Ethics in Natural Language Processing}, pages 53--59, Valencia, Spain. Association for Computational Linguistics.

\bibitem[{Tomsett et~al.(2020)Tomsett, Harborne, Chakraborty, Gurram, and Preece}]{tomsett2020sanity}
Richard Tomsett, Dan Harborne, Supriyo Chakraborty, Prudhvi Gurram, and Alun Preece. 2020.
\newblock Sanity checks for saliency metrics.
\newblock In \emph{Proceedings of the AAAI conference on artificial intelligence}, volume~34, pages 6021--6029.

\bibitem[{Trinh and Mandel(2020)}]{trinh2020directly}
Viet~Anh Trinh and Michael Mandel. 2020.
\newblock Directly comparing the listening strategies of humans and machines.
\newblock \emph{IEEE/ACM Transactions on Audio, Speech, and Language Processing}, 29:312--323.

\bibitem[{Vafa et~al.(2021)Vafa, Deng, Blei, and Rush}]{vafa-etal-2021-rationales}
Keyon Vafa, Yuntian Deng, David Blei, and Alexander Rush. 2021.
\newblock \href {https://doi.org/10.18653/v1/2021.emnlp-main.807} {Rationales for sequential predictions}.
\newblock In \emph{Proceedings of the 2021 Conference on Empirical Methods in Natural Language Processing}, pages 10314--10332, Online and Punta Cana, Dominican Republic. Association for Computational Linguistics.

\bibitem[{Villas-B{\^o}as et~al.(2021)Villas-B{\^o}as, Schwarz, Fontanari, Costa, Cardoso~da Silva, Schneider, Cielo, Spritzer, and Rodrigues~Lobato}]{villas2021acoustic}
Anna~Paula Villas-B{\^o}as, Karine Schwarz, Anna Martha~Vaitses Fontanari, Angelo~Brandelli Costa, Dhiordan Cardoso~da Silva, Maiko~Abel Schneider, Carla~Aparecida Cielo, Poli~Mara Spritzer, and Maria~In{\^e}s Rodrigues~Lobato. 2021.
\newblock Acoustic measures of brazilian transgender women's voices: a case--control study.
\newblock \emph{Frontiers in Psychology}, 12:622526.

\bibitem[{Wang et~al.(2021)Wang, Riviere, Lee, Wu, Talnikar, Haziza, Williamson, Pino, and Dupoux}]{wang-etal-2021-voxpopuli}
Changhan Wang, Morgane Riviere, Ann Lee, Anne Wu, Chaitanya Talnikar, Daniel Haziza, Mary Williamson, Juan Pino, and Emmanuel Dupoux. 2021.
\newblock \href {https://doi.org/10.18653/v1/2021.acl-long.80} {{V}ox{P}opuli: A large-scale multilingual speech corpus for representation learning, semi-supervised learning and interpretation}.
\newblock In \emph{Proceedings of the 59th Annual Meeting of the Association for Computational Linguistics and the 11th International Joint Conference on Natural Language Processing (Volume 1: Long Papers)}, pages 993--1003, Online. Association for Computational Linguistics.

\bibitem[{Wang et~al.(2020{\natexlab{a}})Wang, Tang, Ma, Wu, Okhonko, and Pino}]{wang-etal-2020-fairseq}
Changhan Wang, Yun Tang, Xutai Ma, Anne Wu, Dmytro Okhonko, and Juan Pino. 2020{\natexlab{a}}.
\newblock \href {https://doi.org/10.18653/v1/2020.aacl-demo.6} {Fairseq {S}2{T}: Fast speech-to-text modeling with fairseq}.
\newblock In \emph{Proceedings of the 1st Conference of the Asia-Pacific Chapter of the Association for Computational Linguistics and the 10th International Joint Conference on Natural Language Processing: System Demonstrations}, pages 33--39, Suzhou, China. Association for Computational Linguistics.

\bibitem[{Wang et~al.(2020{\natexlab{b}})Wang, Wu, and Pino}]{wang2020covost}
Changhan Wang, Anne Wu, and Juan Pino. 2020{\natexlab{b}}.
\newblock \href {https://arxiv.org/abs/2007.10310} {Covost 2: A massively multilingual speech-to-text translation corpus}.
\newblock \emph{Preprint}, arXiv:2007.10310.

\bibitem[{Wu et~al.(2023)Wu, Bell, and Rajan}]{wu2023explanations}
Xiaoliang Wu, Peter Bell, and Ajitha Rajan. 2023.
\newblock \href {https://doi.org/10.1109/ICASSP49357.2023.10094635} {Explanations for automatic speech recognition}.
\newblock In \emph{ICASSP 2023 - 2023 IEEE International Conference on Acoustics, Speech and Signal Processing (ICASSP)}, pages 1--5.

\bibitem[{Wu et~al.(2024)Wu, Bell, and Rajan}]{wu2024can}
Xiaoliang Wu, Peter Bell, and Ajitha Rajan. 2024.
\newblock Can we trust explainable ai methods on asr? an evaluation on phoneme recognition.
\newblock In \emph{ICASSP 2024-2024 IEEE International Conference on Acoustics, Speech and Signal Processing (ICASSP)}, pages 10296--10300. IEEE.

\bibitem[{Yan et~al.(2023)Yan, Dalmia, Higuchi, Neubig, Metze, Black, and Watanabe}]{yan-etal-2023-ctc}
Brian Yan, Siddharth Dalmia, Yosuke Higuchi, Graham Neubig, Florian Metze, Alan~W Black, and Shinji Watanabe. 2023.
\newblock \href {https://doi.org/10.18653/v1/2023.eacl-main.119} {{CTC} alignments improve autoregressive translation}.
\newblock In \emph{Proceedings of the 17th Conference of the European Chapter of the Association for Computational Linguistics}, pages 1623--1639, Dubrovnik, Croatia. Association for Computational Linguistics.

\bibitem[{Yin and Neubig(2022)}]{yin-neubig-2022-interpreting}
Kayo Yin and Graham Neubig. 2022.
\newblock \href {https://doi.org/10.18653/v1/2022.emnlp-main.14} {Interpreting language models with contrastive explanations}.
\newblock In \emph{Proceedings of the 2022 Conference on Empirical Methods in Natural Language Processing}, pages 184--198, Abu Dhabi, United Arab Emirates. Association for Computational Linguistics.

\bibitem[{Yu et~al.(2024)Yu, Liu, Ding, Chen, Tao, and Zhang}]{yu-etal-2024-speech}
Tengfei Yu, Xuebo Liu, Liang Ding, Kehai Chen, Dacheng Tao, and Min Zhang. 2024.
\newblock \href {https://doi.org/10.18653/v1/2024.acl-long.435} {Speech sense disambiguation: Tackling homophone ambiguity in end-to-end speech translation}.
\newblock In \emph{Proceedings of the 62nd Annual Meeting of the Association for Computational Linguistics (Volume 1: Long Papers)}, pages 8020--8035, Bangkok, Thailand. Association for Computational Linguistics.

\bibitem[{Zimman(2020)}]{10.1093/oxfordhb/9780190212926.013.45}
Lal Zimman. 2020.
\newblock \href {https://doi.org/10.1093/oxfordhb/9780190212926.013.45} {Transgender language, transgender moment: Toward a trans linguistics}.
\newblock In \emph{The Oxford Handbook of Language and Sexuality}. Oxford University Press.

\end{thebibliography}

\appendix

\section{Occlusion-based Feature Attribution Configuration}
\label{sec:spes-config}

Following the recommendations in \citet{fucci2024spes}, we configured our occlusion-based feature attribution experiments using SLIC segmentation \cite{achanta2012slic} with dynamic adjustment of segments based on input duration. To manage computational costs while maintaining explanation quality for longer inputs, we applied a threshold of 750 frames before capping the segment count. We employ a three-level segmentation strategy by dividing the spectrogram into 2000, 2500, and 3000 segments to obtain segmentations of varying granularity. We generate 20,000 random perturbation masks generated per sample. Each segment has a 0.5 probability of being masked during perturbation. The SLIC sigma parameter was set to 0.

\section{Model Details}
\label{sec:models}

The ST Transformer encoder-decoder model by \cite{wang-etal-2020-fairseq} is a multilingual model trained on the 8 language directions of MuST-C \citep{cattoni2021must} with 72M of parameters, distributed under the MIT License. The model takes as input 80 Mel-filterbank audio features extracted every 10 milliseconds, employing a sample window of 25. The input features are then preprocessed with two 1D convolutional layers with stride 2, reducing input length by a factor of 4, before being fed to the Transformer encoder. We choose this model for our experiments both for its permissive license and its strong gender translation accuracy (see Table \ref{tab:gender-accuracies}).
For inference, we used a beam size of 5 and a no-repeat-ngram-size of 5. 
The maximum source position was set to 7,000.

\begin{table*}[t!]
    \centering
    \setlength{\tabcolsep}{4.5pt}
    \footnotesize
    \begin{tabular}{l||cc|cc|cc|cc|cc|cc}
        \toprule
        & \multicolumn{4}{c|}{\textbf{en-it}} & \multicolumn{4}{c|}{\textbf{en-fr}} & \multicolumn{4}{c}{\textbf{en-es}} \\
        \cmidrule(lr){2-5} \cmidrule(lr){5-9} \cmidrule(lr){9-13}
        & \multicolumn{2}{c|}{\textbf{Cov.}} & \multicolumn{2}{c|}{\textbf{Acc.}} & \multicolumn{2}{c|}{\textbf{Cov.}} & \multicolumn{2}{c|}{\textbf{Acc.}} & \multicolumn{2}{c|}{\textbf{Cov.}} & \multicolumn{2}{c}{\textbf{Acc.}} \\
        \cmidrule(lr){2-13}
        & \textbf{F} & \textbf{M} & \textbf{F} & \textbf{M} & \textbf{F} & \textbf{M} & \textbf{F} & \textbf{M} & \textbf{F} & \textbf{M} & \textbf{F} & \textbf{M} \\
        \midrule
        \textbf{Mult. Transformer} & 54.61 & 53.73 & 77.09 & 94.40 & 53.30 & 54.39 & 77.35 & 91.85 & 66.33 & 65.55 & 80.60 & 91.41 \\
        \textbf{Mono. Conformer} & 49.63 & 46.99 & 46.38 & 75.76 & 56.84 & 56.10 & 49.80 & 72.50 & 69.13 & 65.07 & 39.21 & 76.74 \\
        \textbf{Large-scale Conformer} & - & - & - & - & - & - & - & - & 76.28 & 71.53 & 53.27 & 69.52 \\
        \bottomrule
    \end{tabular}
    \caption{Coverage (Cov.) and Accuracy (Acc.) for the the models considered across the three language pairs.}
    \label{tab:gender-accuracies}
\end{table*}

\section{Dataset Description and Processing}
\label{sec:data}

Our analysis uses MuST-SHE \cite{bentivogli-etal-2020-gender}, a speech translation dataset compiled from TED talks that includes annotations for words which are gender-neutral in the source language but have gender marking in the reference translation to the target language. The dataset is licensed under CC BY NC ND 4.0 International, and our usage aligns with its intended purpose of studying gender term translation in speech translation systems. Comprehensive information about the dataset's domain coverage, language scope, linguistic phenomena, and represented demographic groups is available in the dataset's data statement.\footnote{https://mt.fbk.eu/data-statement-for-must-she/} Given its origin in educational talks, the corpus is free of offensive content. It contains identifiable information in the form of speaker names, which are publicly available with their consent as part of their TED talks. The dataset includes speaker's gender information, which was annotated based on speakers' self-identification as documented in their public profiles at the time of dataset creation.

We run our experiments on English-to-Spanish/French/Italian translations, specifically examining gender terms that refer to the speaker (category 1 in MuST-SHE), as these instances may benefit from acoustic cues for gender disambiguation. The dataset provides pairs of correct and incorrect gender translations for each term (e.g., `curiosa'/`curioso'), which we use in our contrastive analysis. Using the annotations from \citet{savoldi-etal-2022-morphosyntactic}, we exclude gender articles from our analysis due to their high frequency in both genders across sentences, which makes it challenging to reliably identify specific instances referring to the speaker and could introduce noise into the evaluation.

Our analysis encompasses only those gender terms for which the model generates one of the forms annotated in MuST-SHE (either correct or incorrect version), as these are the only instances where we have access to the contrastive alternative. For the Transformer model on English-to-French translation, for which we present experimental results in the main paper, this filtering process yields a final set of 975 terms for analysis.

\section{Full Deletion Range Analysis}
\label{sec:no-zoom}

\begin{figure}[ht]
    \centering
    \includegraphics[width=\columnwidth]{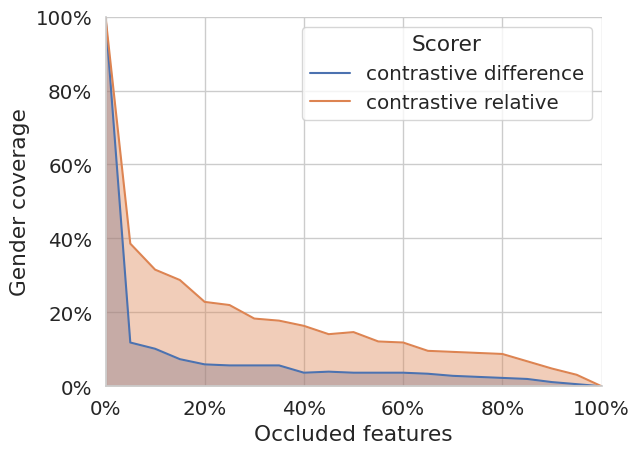}
    \caption{Gender coverage for all deletion steps for explanations with different scorers for the Transformer model on en-fr data.}
    \label{fig:cov-no-zoom}
\end{figure}

\begin{figure}[ht]
    \centering
    \includegraphics[width=\columnwidth]{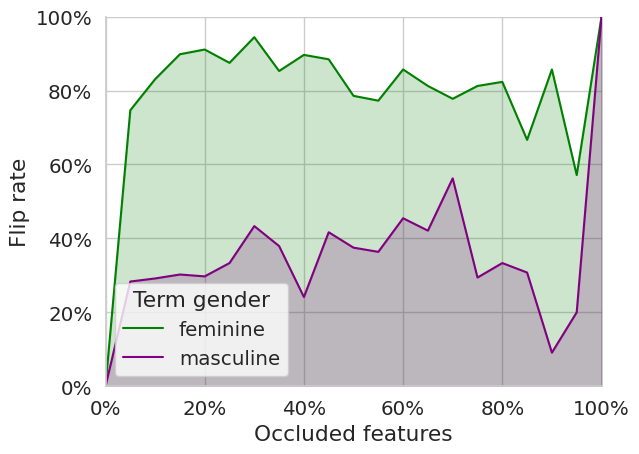}
    \caption{Flip rate for all deletion steps for explanations with different scorers for the Transformer model on en-fr data.}
    \label{fig:flip-no-zoom}
\end{figure}

The choice to focus on the first 20\% of feature deletion in our main analysis stems from the fact that beyond this point the input becomes too degraded for model outputs to be meaningful. As shown in Figure \ref{fig:cov-no-zoom}, coverage drops substantially with increased deletion, making the flip rate measurements unreliable, as evidenced by the erratic behavior of the curve in Figure \ref{fig:flip-no-zoom} at higher deletion percentages.

\section{Extended Results Across Models and Languages}
\label{sec:other-experiments}

In this section, to make sure our contrastive explanation method works for different models and language pairs, we extend our study to the English-Italian and English-Spanish sections of the data (\S\ref{sec:additional_langs}) and to two other types of models. In \S\ref{sec:additional_conformer}, we analyze the monolingual Conformer encoder-Transformer decoder models by \citet{papi-etal-2024-good}. Similarly to the Transformer model used in the main body of the paper, this model is trained on MuST-C. In \S\ref{sec:additional_largescale}, to analyze a larger-scale system, we study a Conformer encoder-Transformer decoder model trained on the English-Spanish data of the constrained track of the last IWSLT campaign \citep{ahmad-etal-2024-findings}. 
While our perturbation-based method for contrastive explanations is inherently model-agnostic and applicable to any architecture, we empirically validate that the explanations generated across these diverse model configurations continue to satisfy our quantitative faithfulness metrics.

\subsection{Results on Additional Language Pairs}
\label{sec:additional_langs}

\begin{figure}[t]
    \centering
    \begin{subfigure}[t]{0.47\textwidth}
            \centering
            \label{fig:transformer-it-coverage}
            \includegraphics[width=\linewidth,  keepaspectratio]{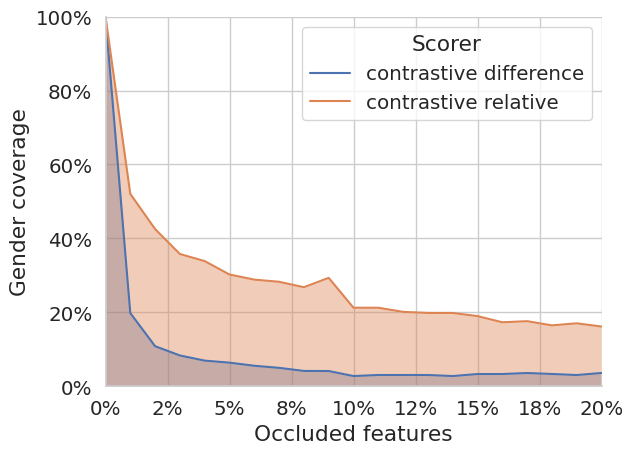}
            \caption{en-it}
        \end{subfigure}
        \hfill
        \begin{subfigure}[t]{0.47\textwidth}
            \centering
            \label{fig:transformer-es-coverage}
            \includegraphics[width=\linewidth, keepaspectratio]{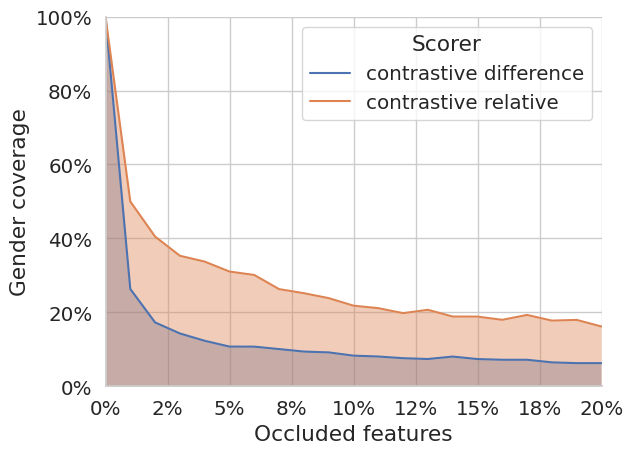}
            \caption{en-es}
        \end{subfigure}
    \caption{Coverage for the first 20\% deletion steps for explanations with the difference and relative scorers for the Transformer model on English-Italian and English-Spanish.}
    \label{fig:transformer-coverage}
\end{figure}

For the multilingual Transformer model, we extend our analysis to English-Italian and English-Spanish translations to verify whether the patterns observed for English-French generalize across language pairs. As shown in 
Figure \ref{fig:transformer-coverage},
the relative scorer maintains substantially higher coverage than the difference scorer across 
the two languages, confirming the findings reported in the main paper for English-French.
When deleting 10\% of the features, coverage remains above 20\% for Italian and Spanish, matching the robustness observed for French translations.

\begin{figure}[t]
    \centering
    \begin{subfigure}[t]{0.47\textwidth}
            \centering
            \label{fig:transformer-it-flip}
            \includegraphics[width=\linewidth,  keepaspectratio]{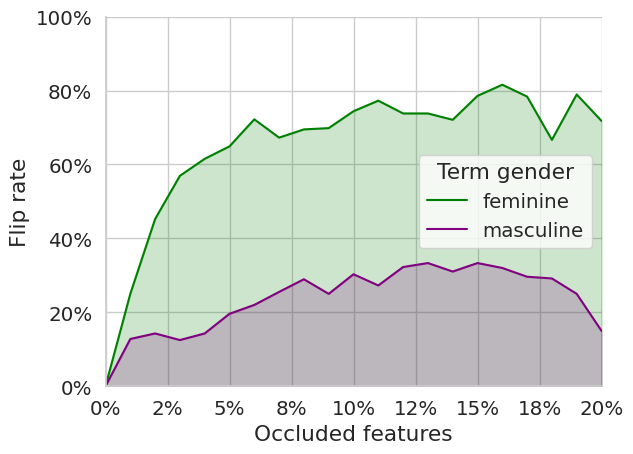}
            \caption{en-it}
        \end{subfigure}
        \hfill
        \begin{subfigure}[t]{0.47\textwidth}
            \centering
            \label{fig:transformer-es-flip}
            \includegraphics[width=\linewidth, keepaspectratio]{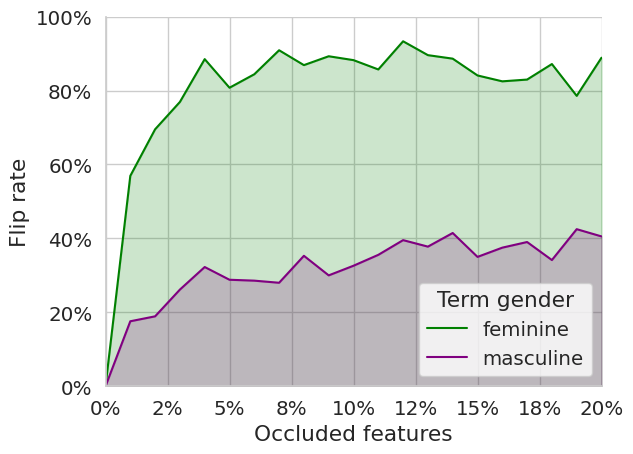}
            \caption{en-es}
        \end{subfigure}
    \caption{Flip rate for the first 20\% deletion steps for explanations with the relative scorer for the Transformer model on English-Italian and English-Spanish.}
    \label{fig:transformer-flip}
\end{figure}

The gender-flipping behavior is also consistent across languages, as shown in Figure \ref{fig:transformer-flip}. For feminine predictions,
deleting just 5\% of the most relevant features identified by our method causes the model to switch to masculine translations in over 60\% of the covered cases
for both Italian and Spanish, as observed for French in \S\ref{sec:use-case}.
The asymmetry between feminine and masculine predictions persists as well, with masculine-to-feminine flip rates not going much higher than 40\% across all language pairs. This consistent pattern across three Romance languages with similar grammatical gender systems suggests that our earlier hypothesis about a masculine default could plausibly explain the model's behavior in these three cases.

The consistency across three target languages shows the robustness of our method and suggests that the model may be using similar strategies to make gender choices in all languages considered.

\subsection{Results on Conformer Models}
\label{sec:additional_conformer}

In this section, we report the results obtained for the three language pairs en-es/fr/it using the monolingual Conformer encoder-Transformer decoder models by \citet{papi-etal-2024-good}, released under the Apache 2.0 License. Such models feature a higher translation quality than the Transformer-based model we reported on above, but has a lower gender accuracy, as reported in Table \ref{tab:gender-accuracies}.

\begin{figure}
    \centering
        \begin{subfigure}[t]{\columnwidth}
            \centering
            \label{fig:conformer-es-coverage}
            \includegraphics[width=\linewidth, keepaspectratio]{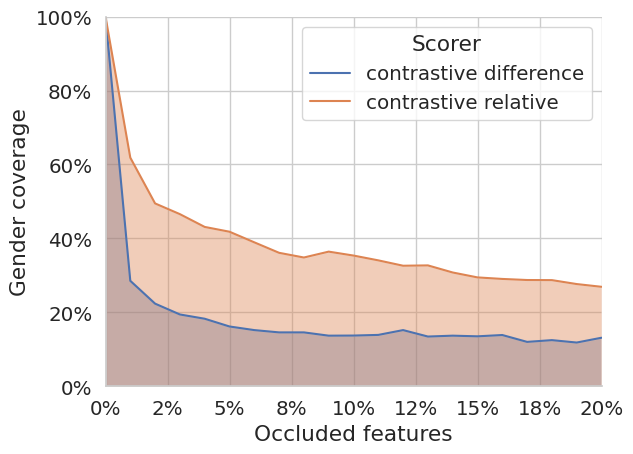}
            \caption{en-es}
        \end{subfigure}
        \begin{subfigure}[t]{\columnwidth}
            \centering
            \label{fig:conformer-fr-coverage}
            \includegraphics[width=\linewidth, keepaspectratio]{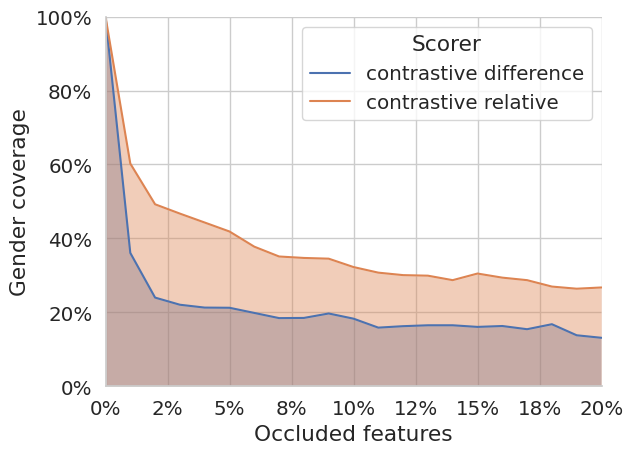}
            \caption{en-fr}
        \end{subfigure}
          \hfill      
    \begin{subfigure}[t]{\columnwidth}
            \centering
            \label{fig:conformer-it-coverage}
            \includegraphics[width=\linewidth,  keepaspectratio]{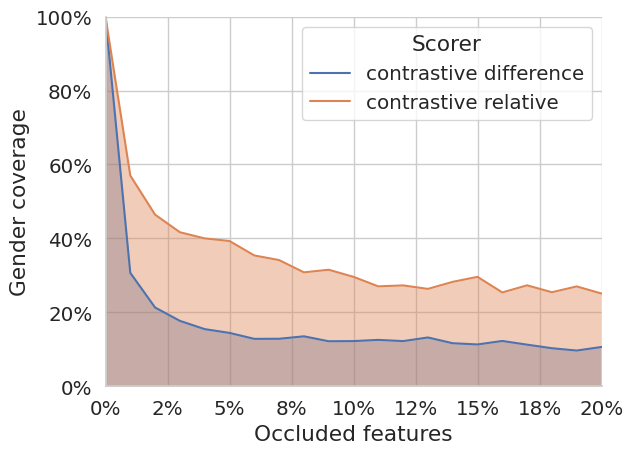}
            \caption{en-it}
        \end{subfigure}

    \caption{Coverage for the first 20\% deletion steps for explanations with the difference and relative scorers for the Conformer models on different language pairs.}
    \label{fig:conformer-coverage}
\end{figure}

The coverage patterns observed for the Transformer model extend to the monolingual Conformer models. As shown in Figure \ref{fig:conformer-coverage}, the relative scorer maintains consistently higher coverage than the difference scorer, although with lower absolute values compared to the Transformer model. At 10\% deletion, coverage with the relative scorer remains above 30\% for all language pairs, indicating that our method still effectively isolates gender-relevant features despite the architectural differences.

\begin{figure}
    \centering
        \begin{subfigure}[t]{\columnwidth}
            \centering
            \label{fig:conformer-es-flip}
            \includegraphics[width=\linewidth, keepaspectratio]{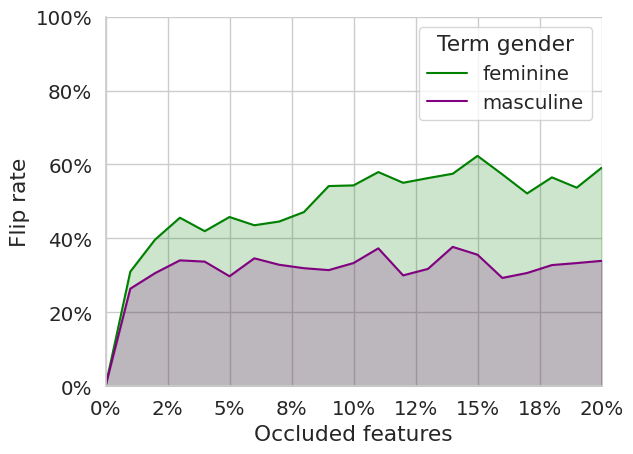}
            \caption{en-es}
        \end{subfigure}
        \begin{subfigure}[t]{\columnwidth}
            \centering
            \label{fig:conformer-fr-flip}
            \includegraphics[width=\linewidth, keepaspectratio]{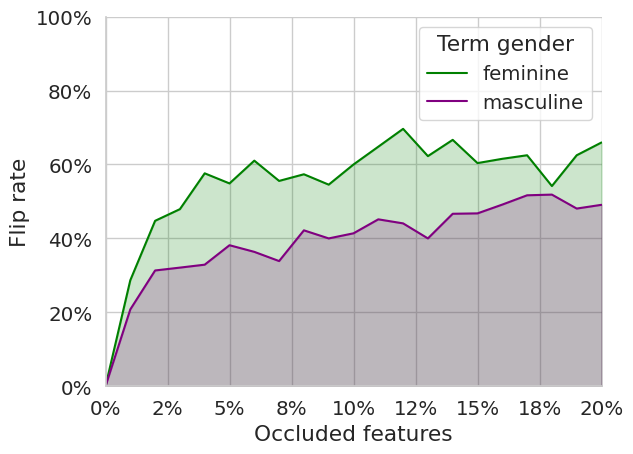}
            \caption{en-fr}
        \end{subfigure}    
    \begin{subfigure}[t]{\columnwidth}
            \centering
            \label{fig:conformer-it-flip}
            \includegraphics[width=\linewidth,  keepaspectratio]{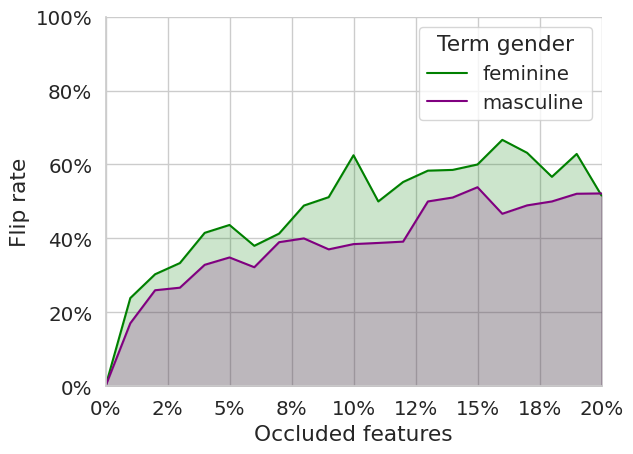}
            \caption{en-it}
        \end{subfigure}

    \caption{Flip rate for the first 20\% deletion steps for explanations with the relative scorer for the Conformer models on different language pairs.}
    \label{fig:conformer-flip}
\end{figure}

The gender-flipping behavior shows interesting variations from the Transformer results. As seen in Figure \ref{fig:conformer-flip}, while the relative scorer still triggers gender switches, the effect is more balanced between feminine and masculine predictions - both achieving flip rates between 30-60\% when 10\% of features are deleted. This differs from the Transformer's strong asymmetry between genders. This more 
balanced gender-flipping
effect aligns with the Conformer models' general gender translation patterns (Table \ref{tab:gender-accuracies}). Indeed, these models show substantially lower gender accuracy, particularly for feminine terms referring to the speaker, where performance approaches chance level. The reduced impact of feature deletion on gender prediction suggests these models 
may rely not only on the speech input but also on other factors, such as previously generated tokens or 
potential biases embedded in the decoder's ``internal language model'' 
\cite{fucci-etal-2023-integrating}.

\subsection{Results on Large-scale Model}
\label{sec:additional_largescale}

To further confirm the robustness of our findings, we investigate whether they also hold for larger-scale models trained on different data from only MuST-C. To this aim, we train a model composed of a 12-layer Conformer encoder and 6-layer Transformer decoder. Besides MuST-C, the training data includes EuroParl-ST \citep{europarlst}, CoVoST v2 \citep{wang2020covost}, and the ASR datasets CommonVoice \citep{ardila-etal-2020-common}, LibriSpeech \citep{librispeech}, TEDLIUM v3 \citep{tedlium}, and VoxPopuli \citep{wang-etal-2021-voxpopuli}, whose transcripts we automatically translate into Spanish using the NeMo 
MT models.\footnote{Publicly available at: \url{https://docs.nvidia.com/deeplearning/nemo/user-guide/docs/en/main/nlp/machine_translation/machine_translation.html}.} The model is trained with a composite loss function that comprises a label-smoothed cross entropy on the decoder output, with the translation as target, and two auxiliary CTC losses, respectively, on the encoder output, with the translation as target and, on the 8th encoder layer, with the transcript as target \citep{yan-etal-2023-ctc}. The training is performed using the Noam scheduler with 2e-3 as peak learning rate and is stopped after 200,000 updates. Utterance-level Cepstral Mean and Variance Normalization (CMVN) and SpecAugment \cite{park2019specaugment} are applied during training and segments longer than 30 seconds are filtered out (fairseq-ST default) to avoid excessive VRAM requirements. The model is the average of the last 7 checkpoints obtained from the training and has 133M parameters. The training is run on 4 NVIDIA Ampere GPU A100 (64GB VRAM) with mini-batches of 40,000 input elements and 2 as update frequency.

\begin{figure}[ht]
    \centering
    \includegraphics[width=\columnwidth]{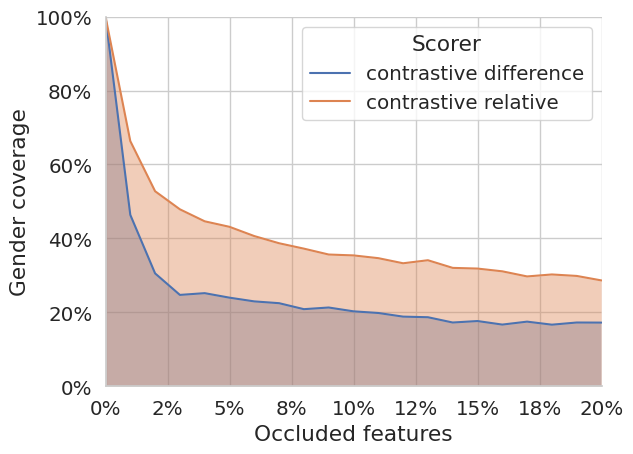}
    \caption{Coverage for the first 20\% deletion steps for explanations with the difference and relative scorers for the large-scale Conformer model on English-Spanish.}
    \label{fig:big-conformer-es-coverage}
\end{figure}

\begin{figure}[ht]
    \centering
    \includegraphics[width=\columnwidth]{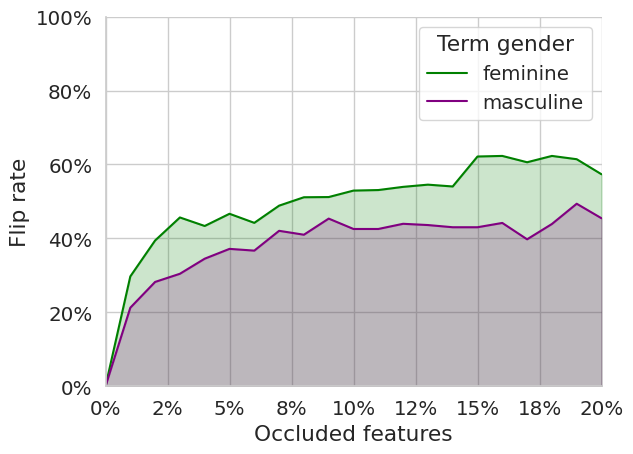}
        \caption{Flip rate for the first 20\% deletion steps for explanations with the relative scorer for the large-scale Conformer models on English-Spanish.}

    \label{fig:big-conformer-es-flip}
\end{figure}

This large-scale model exhibits similar patterns to the smaller Conformer models. As shown in Figures \ref{fig:big-conformer-es-coverage} and \ref{fig:big-conformer-es-flip}, the relative scorer maintains higher coverage than the difference scorer. The flip rate displays a reduced asymmetry between feminine and masculine predictions, with the first reaching around 50\% after 10\% feature deletion, and the second, a little over 40\%. This behavior is aligned with that of the monolignual small Conformer models.

\section{Comparison of Subword-to-Word Probability Aggregation Techniques}
\label{sec:word-probs}

\begin{table*}[t]
\centering
\begin{tabular}{llll}
\hline
\textbf{Method Comparison} & \textbf{en-fr} & \textbf{en-es} & \textbf{en-it} \\
\hline
Word Boundary vs. Length Norm & $<$ 0.001*** & 0.218 & $<$ 0.001*** \\
Word Boundary vs. Chain Rule & 0.376 & $<$ 0.001*** & $<$ 0.001*** \\
Length Norm vs. Chain Rule & $<$ 0.001*** & $<$ 0.01** & $<$ 0.001*** \\
\hline
\end{tabular}
\begin{minipage}{\textwidth}
\centering
\vspace{0.5em}
*** p $<$ 0.001, ** p $<$ 0.01
\caption{Statistical significance (p-values) of pairwise method comparisons for coverage performance across language pairs.}
\label{tab:stats-cov}
\end{minipage}
\end{table*}

\begin{table*}[t]
\centering
\begin{tabular}{llll}
\hline
\textbf{Method Comparison} & \textbf{en-fr} & \textbf{en-es} & \textbf{en-it} \\
\hline
Word Boundary vs. Length Norm & $<$ 0.001*** & 0.196 & 0.129 \\
Word Boundary vs. Chain Rule & 0.605 & 0.998 & 0.657 \\
Length Norm vs. Chain Rule & $<$ 0.001*** & 0.110 & 0.074 \\
\hline
\end{tabular}
\begin{minipage}{\textwidth}
\centering
\vspace{0.5em}
*** p $<$ 0.001
\caption{Statistical significance (p-values) of pairwise method comparisons for flip rate performance across language pairs.}
\label{tab:stats-flip-rate}
\end{minipage}
\end{table*}

This section presents an empirical comparison, for our XAI application, of different methods for computing word-level probabilities from the sequences of subword token probabilities output by ST models. Previous work has employed various approaches to address this issue. Some studies have avoided the problem entirely by limiting their analysis to targets and foils tokenized as single tokens \cite{yin-neubig-2022-interpreting}. Others have worked at the subword level without aggregation \cite{ferrando-etal-2022-towards}, or employed the chain rule to multiply individual token probabilities \cite{sarti-etal-2023-inseq}. Here, we leverage our evaluation metrics for explanation faithfulness---coverage and flip rate---to compare three methods and determine if any of them yields more reliable contrastive explanations when integrated into our methodology.

The \textbf{Chain Rule} method simply multiplies the probabilities of all subword tokens that compose a word, applying the standard chain rule of probability:
\begin{equation}
p(w) = \prod_{i=0}^{n-1} p(w_i)
\end{equation}
This approach is commonly used for aggregating subword probabilities, and serves as the default method in \cite{sarti-etal-2023-inseq}. While straightforward, this method does not account for the varying number of tokens across different words, potentially biasing toward shorter token sequences.

The \textbf{Length Normalization} method addresses this limitation by taking the n\textsuperscript{th} root of the product of token probabilities, where n is the number of tokens:
\begin{equation}
p(w) = \sqrt[n]{\prod_{i=0}^{n-1} p(w_i)}
\end{equation}
In the log domain, this corresponds to computing the average log probability over the sequence of subwords, effectively normalizing by the sequence length. This normalization, which is typically used in beam search, attempts to create a fairer comparison between words of different lengths \cite{meister-etal-2020-beam}.

The \textbf{Word Boundary} method, proposed by \citet{pimentel-meister-2024-compute}, considers not only the probability of the subword sequence itself but also the probability that this sequence forms a complete word rather than being part of a longer word. This is calculated using:
\begin{equation}
p(w)=\prod_{i=0}^{n-1} p(w_i) \cdot \frac{p(S^{bow}_{n+1})}{p(S^{bow}_0)}
\end{equation}
where $S^{bow}_i$ is the set of beginning-of-word tokens at step $i$. This approach ensures that the subwords $w_{0,...,n} \in w$ are followed by a new word boundary, preventing probability overestimation for words that are prefixes of longer terms.

\begin{figure}
    \centering
        \begin{subfigure}[t]{\columnwidth}
            \centering
            \includegraphics[width=\linewidth, keepaspectratio]{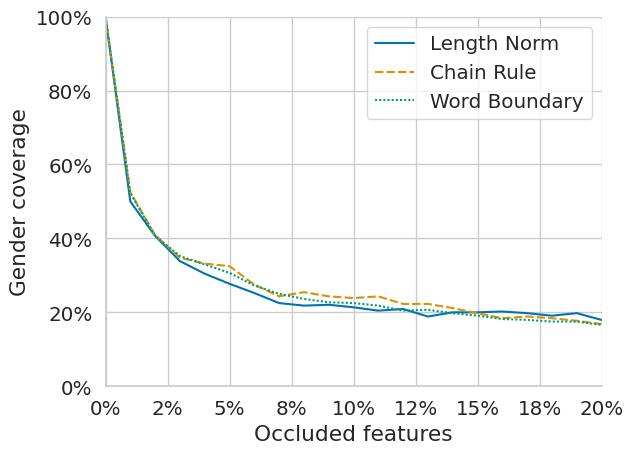}
            \caption{en-es}
        \end{subfigure}
        \begin{subfigure}[t]{\columnwidth}
            \centering
            \includegraphics[width=\linewidth, keepaspectratio]{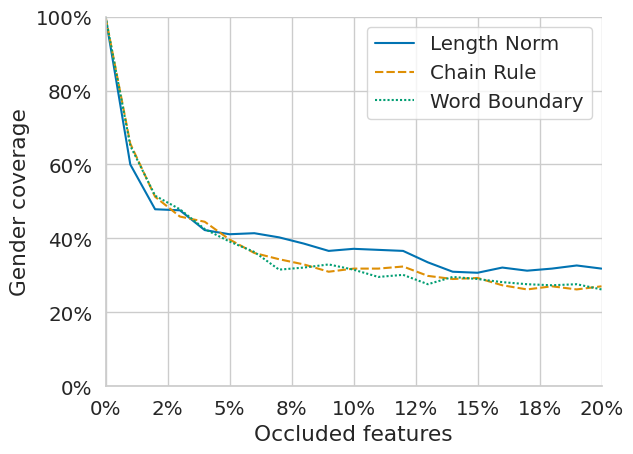}
            \caption{en-fr}
        \end{subfigure}
          \hfill      
    \begin{subfigure}[t]{\columnwidth}
            \centering
            \includegraphics[width=\linewidth,  keepaspectratio]{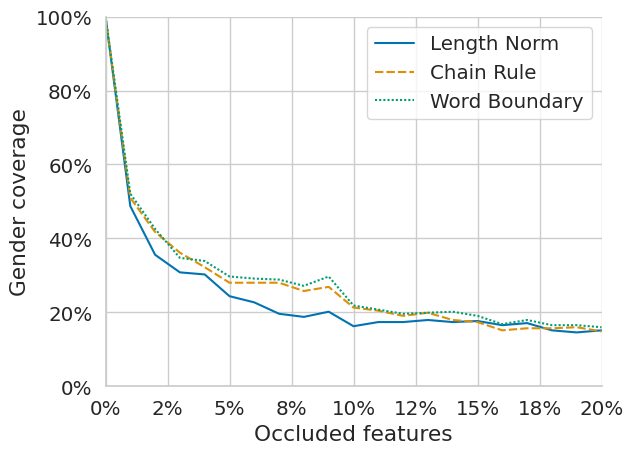}
            \caption{en-it}
        \end{subfigure}
    \caption{Coverage for the first 20\% deletion steps for explanations obtained with different word probability computation methods, the Transformer model, and different language pairs.}
    \label{fig:word-probs-coverage}
\end{figure}

\begin{figure}
    \centering
        \begin{subfigure}[t]{\columnwidth}
            \centering
            \includegraphics[width=\linewidth, keepaspectratio]{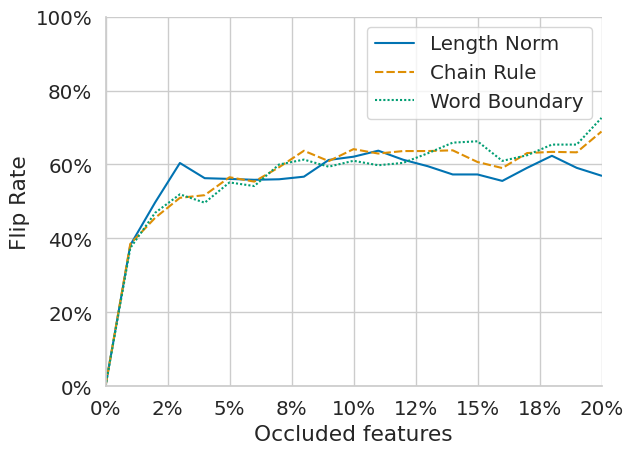}
            \caption{en-es}
        \end{subfigure}
        \begin{subfigure}[t]{\columnwidth}
            \centering
            \includegraphics[width=\linewidth, keepaspectratio]{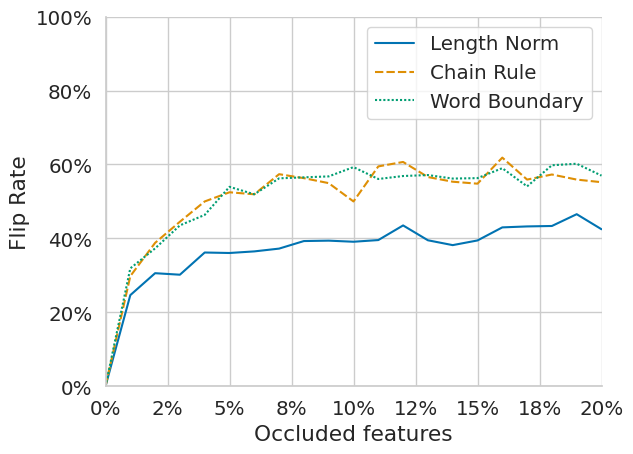}
            \caption{en-fr}
        \end{subfigure}
          \hfill      
    \begin{subfigure}[t]{\columnwidth}
            \centering
            \includegraphics[width=\linewidth,  keepaspectratio]{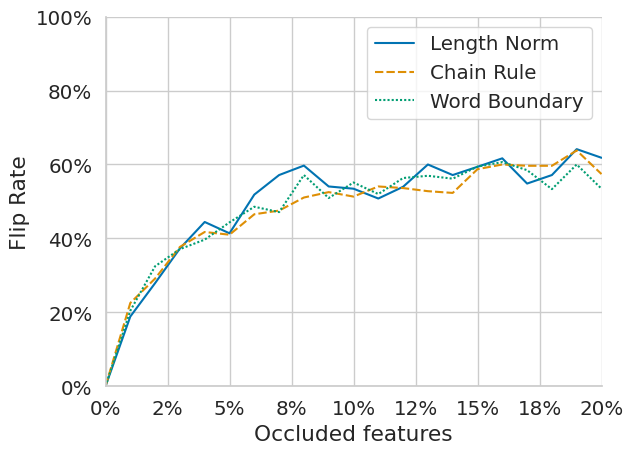}
            \caption{en-it}
        \end{subfigure}
    \caption{Flip rate for the first 20\% deletion steps for explanations obtained with different word probability computation methods, the Transformer model, and different language pairs.}
    \label{fig:word-probs-flip-rate}
\end{figure}

\begin{table*}
    \small
    \centering
    \begin{tabular}{l|ccc|cc|c}
    \toprule
     & \multicolumn{3}{c|}{\textbf{Mult. Transformer}} & \multicolumn{2}{c|}{\textbf{Mono. Conformer}} & \textbf{Large-scale Conformer} \\
    
     & \textbf{en-it} & \textbf{en-fr} & \textbf{en-es} & \textbf{en-it} & \textbf{en-fr} & \textbf{en-es} \\
    \midrule
    \textbf{Difference} & 0.94 & 0.93 & 0.92 & 0.88 & 0.86 & 0.89 \\
    \textbf{Relative} & 0.36 & 0.33 & 0.39 & 0.29 & 0.32 & 0.43 \\
    \bottomrule
    \end{tabular}
    \caption{Pearson correlation between non-contrastive explanations and contrastive explanations obtained with the difference and relative scorer for different models and language pairs.}
  \label{tab:corr}
\end{table*}

To evaluate the performance of each method, we recorded coverage and flip rate measurements at progressive deletion steps, as shown in Figures \ref{fig:word-probs-coverage} and \ref{fig:word-probs-flip-rate}. To quantitatively assess differences between methods, we conducted paired t-tests comparing coverage and flip rate values. Each pair of methods was evaluated using the sequence of measurements at corresponding deletion percentages as dependent samples, allowing us to determine whether observed differences were statistically significant. Tables \ref{tab:stats-cov} and \ref{tab:stats-flip-rate} present the resulting p-values for each comparison across language pairs.

These comparisons reveal that all three methods perform similarly in their ability to identify gender-relevant features. While coverage differences  are often statistically significant (Table \ref{tab:stats-cov}), the absolute differences are relatively small as can be seen in Figure \ref{fig:word-probs-coverage}, with all methods maintaining sufficient coverage levels to ensure reliable flip rate calculations. The flip rate analysis---our primary metric for evaluating how precisely methods isolate features responsible for gender prediction---shows no statistically significant differences ($p > 0.05$) between Word Boundary and Chain Rule methods across all language pairs (Table \ref{tab:stats-flip-rate}). As we see in Figure \ref{fig:word-probs-flip-rate}, when over 10\% of the highlighted features are occluded, all methods achieve comparable flip rates between 50-60\% for most language pairs. The only notable exception is the Length Normalization method, which performs significantly worse for French in terms of flip rate (Table \ref{tab:stats-flip-rate}, $p < 0.001$). 
While the reasons for this language-specific underperformance warrant further investigation, the observation supports our decision to exclude the Length Normalization method from our main experiments.

Since the Chain Rule and Word Boundary methods yield comparable performance, we looked at individual examples in which the ranking of the target and foil is different with the two methods to deepen our comparison. Namely, we found a few examples in which the Chain Rule method assigns a higher probability to the foil than to the target, even though the beam search selects the target word in the end.
For instance, in the sentence ``In one, I was the classic Asian \underline{student}, relentless in the demands that I made on myself'', translated into Italian as ``In uno, ero la classica \underline{studentessa} asiatica, incantata nelle richieste di me stessa'', the feminine form \textit{studentessa}---tokenized as \texttt{\_studente ssa}---has a joint probability of 0.899, while the masculine form \texttt{\_studente} scores 0.908. 
However, when applying the Word Boundary method, the unchosen masculine form receives a much lower probability (0.004), reflecting a more accurate estimate. This method's principled treatment of complete words versus prefixes makes it especially suited for analyzing gendered terms, where precise probability computation is critical. However, the number of cases in which this happens is fairly limited (less than 10), which explains why there are no significant differences between the scores of the two methods when aggregating over the whole dataset. Nonetheless, this example demonstrates that in the few cases in which they differ, the Word Boundary method provides the most correct probability estimation.

\section{Correlation Between Contrastive and Non-Contrastive Explanations}
\label{sec:correlations}

The patterns regarding similarity between contrastive and non-contrastive explanations observed for the Transformer model on English-French translations extend consistently across all models and language pairs. 
Table \ref{tab:corr} shows the Pearson correlation coefficients between non-contrastive explanations and contrastive explanations generated using the difference and relative scorers, respectively. The difference scorer produces explanations that strongly correlate with non-contrastive ones (correlation coefficients between 0.86 and 0.94), while the relative scorer generates distinctly different explanations (correlation coefficients between 0.29 and 0.43). These results reinforce our findings from Section \ref{sec:use-case} regarding the relative scorer's superior ability to generate truly contrastive explanations.

\end{document}